\documentclass{ieeeaccess}
\usepackage{cite}
\usepackage{amsmath,amssymb,amsfonts}
\usepackage{algorithmic}
\usepackage{graphicx}
\usepackage{textcomp}
\usepackage{subcaption}
\usepackage{subfig}
\usepackage{booktabs}  
\usepackage{tabularx}  
\usepackage{multirow}  
\usepackage{url}
\usepackage{threeparttable}

\def\BibTeX{{\rm B\kern-.05em{\sc i\kern-.025em b}\kern-.08em
    T\kern-.1667em\lower.7ex\hbox{E}\kern-.125emX}}
\begin{document}
\history{Date of publication xxxx 00, 0000, date of current version xxxx 00, 0000.}
\doi{10.1109/ACCESS.2024.DOI}

\title{Investigation of Neural Network Methods for Reconstruction and Classification of Texture Images Under Conditions of Incomplete Information}

\author{
\uppercase{G. Abdimanap}\authorrefmark{1,2,3},
\uppercase{K. Bostanbekov}\authorrefmark{2,3},
\uppercase{A. Abdallah}\authorrefmark{4},
\uppercase{A. Alimova}\authorrefmark{2},
\uppercase{D. Kurmangaliyev}\authorrefmark{2},
\uppercase{D. Nurseitov}\authorrefmark{1,2},
\uppercase{T. Dedova}\authorrefmark{3},
\uppercase{L. Balakay}\authorrefmark{3}
\uppercase{AND S. Nurakynov}\authorrefmark{3}
}

\address[1]{Satbayev University, Almaty, 050013, Kazakhstan}
\address[2]{KazMunayGas Engineering LLP, Astana, 010000, Kazakhstan}
\address[3]{Institute of Ionosphere LLP, Almaty, 050020, Kazakhstan}
\address[4]{Information Technology Department, Assiut University, Assiut, 71515, Egypt}

\tfootnote{This research was funded by the Science Committee of the Ministry of Science and Higher Education of the Republic of Kazakhstan (Grant No. AP26102621)}

\markboth
{Abdimanap \headeretal: Investigation of Neural Network Methods for Reconstruction and Classification}
{Abdimanap \headeretal: Investigation of Neural Network Methods for Reconstruction and Classification}

\corresp{Corresponding author: A. Alimova (e-mail: a.alimova@kmge.kz).}

\begin{abstract}
The automated analysis of heterogeneous natural textures is frequently hindered by 
physical damage and data loss, presenting a significant challenge to computer vision. 
While deep learning has shown success in controlled environments, its application to 
complex geological materials under conditions of incomplete information remains 
underexplored. This study presents an integrated framework for the inpainting and 
classification of high-resolution core sample images. We propose an end-to-end 
pipeline that utilizes object detection for sample segmentation, followed by image 
inpainting using Generative Adversarial Networks (GANs) with Contextual Residual 
Aggregation (CRA) to reconstruct missing high-frequency details. Subsequently, we 
evaluate the performance of modern Transformer-based (Swin, ViT) and Convolutional 
Neural Network (CNN) architectures on the reconstructed data. Our experiments revealed a critical divergence between reconstruction quality 
and downstream utility: despite high structural fidelity (PSNR 28.7~dB, FID 
74.01), classification accuracy plateaued at 53\%. To improve minority-class 
detection, we propose a confidence-based hybrid ensemble that raises MCA from 
48\% to 58\%. These results highlight the 
limitations of current state-of-the-art generative models, which may produce 
visually plausible but semantically ambiguous features ("hallucinations") that 
confound classifiers. This work provides insights into the dependencies between 
image reconstruction quality and classification performance, offering a reproducible baseline for future research in non-destructive testing and
material science. Given that cross-well accuracy remains in the 49--53\% range, we
position the resulting system as a decision-support and screening tool for
lithofacies interpretation rather than as a fully autonomous classifier. The code is available at 
\url{https://github.com/GalymzhanAbdimanap/Lithology_recognition}
\end{abstract}

\begin{keywords}
Deep Learning; Image Inpainting; Texture Classification; Vision Transformers; Generative Adversarial Networks (GAN); Material Analysis
\end{keywords}

\titlepgskip=-15pt

\maketitle

\section{Introduction}
\label{sec:introduction}

The analysis of high-resolution optical and hyperspectral data plays a pivotal role across domains such as sensing, medical imaging, materials science, and geological exploration~\cite{10.5382Geo-and-Mining-09,10.5382rev.21.07,gandhi2016essentials}. A critical bottleneck in these applications is the quality of the input data: captured images often contain artifacts, occlusions, or physical damage, which significantly hinder automated processing. In particular, the geological domain presents a unique computer vision challenge: drill-core samples—cylindrical rock sections used to determine lithology—are characterized by extreme textural heterogeneity and are often damaged during extraction, leading to data loss~\cite{krahenbuhl2015new,202Editorial3}. A recent comprehensive review~\cite{gunther2025machine} highlights that machine learning has become an indispensable tool for automating the analysis of complex drill-core images, addressing challenges ranging from feature extraction to lithological logging.

Traditionally, the processing of such complex visual data relies on manual interpretation, which is subjective and non-scalable~\cite{Resolution2021}. While Deep Learning (DL) has revolutionized image recognition, standard Convolutional Neural Networks (CNNs) and Vision Transformers (ViTs) degrade significantly when applied to incomplete or “inpainted” data where the semantic context is artificially reconstructed. Although techniques like Hyperspectral Imaging (HSI) provide rich spectral information in the visible-near infrared (VNIR) and short-wave infrared (SWIR) ranges~\cite{clark1999spectroscopy,pan2014electron,van2004analysis}, the problem of structural discontinuity (holes and cracks) remains an open research question.

In this work, we investigate the relationship between generative image inpainting and semantic classification in complex, high-entropy scenarios. Using geological core images as a "hard-case" benchmark, our research examines how modern GAN-based models restore missing data and how this reconstruction impacts the reliability of automated classification. Our principal contribution is an integrated processing pipeline coupled with a rigorous empirical analysis demonstrating that high perceptual reconstruction quality does not guarantee downstream classification utility. This analysis explicitly quantifies the semantic gap between reconstruction fidelity and classification performance, and informs a dual-configuration deployment strategy tailored to the operational constraints of field and laboratory 
environments.

The primary contributions of this research are as follows:
\begin{enumerate}
    \item \textbf{Integrated Pipeline for Damaged Data:} Development of a unified workflow combining object detection, instance segmentation, and GAN-based inpainting to systematically process geologically complex images with physical imperfections.
    
    \item \textbf{Decoupling Visual and Semantic Quality:} Investigation of the impact of CRA inpainting on downstream tasks, with an explicit analysis of the trade-off between perceptual fidelity and semantic preservation.

    \item \textbf{Robustness Benchmarking:} Rigorous comparative analysis of CNNs (EfficientNet) and Vision Transformers (Swin) on reconstructed data, assessing their sensitivity to synthetic artifacts introduced by generative models.
    
    \item \textbf{Quantification of the Semantic Gap:} Empirical demonstration that high perceptual reconstruction quality does not guarantee downstream classification utility, and proposal of a confidence-based hybrid ensemble to improve detection of underrepresented lithofacies.
\end{enumerate}

The remainder of this paper is organized as follows: 
Section~\ref{sec:related_work} reviews related work, 
Section~\ref{sec:prosedwork} details the proposed methodology, 
Section~\ref{sec:approach-multi-modal} presents the experimental results 
and ablation study, Section~\ref{sec:discussion} discusses the findings, 
and Section~\ref{sec:conclusion} concludes with future research directions.

\section{Background and Related Work}
\label{sec:related_work}
This section reviews recent advancements in machine-learning-based analysis of drill cores and provides a technical overview of the CNN architectures and image inpainting techniques relevant to this study.

\subsection{Recent Work on Drill-cores}

Recent studies have demonstrated the potential of machine learning for lithofacies classification. \cite{CORINA2018664} proposed an integrated technique combining lithofacies classification with well-log interpretations to model core permeability. Using probabilistic neural networks (PNNs), they achieved an overall accuracy of 95.81\% in predicting discrete lithofacies at missing intervals. Similarly, \cite{HE2019410} employed a multilayer perceptron classifier based on facies analysis and statistical classification, achieving accuracies between 66\% and 99\% depending on the available wireline log data.

Deep learning has further pushed these boundaries. \cite{zhang2017deep} utilized a CNN architecture to determine lithology from borehole image logs, achieving approximately 95\% accuracy. \cite{fu2022deep} demonstrated that automated frameworks could outperform traditional manual logging in both consistency and speed. Beyond basic identification, recent research has expanded into mineralogical analysis; for instance, \cite{boiger2024direct} used transfer learning to predict mineral content directly from core images, capturing subtle visual cues often imperceptible to the human eye. \cite{Caja_10} also showed the efficacy of supervised learning (SVM) in identifying lithology from high-resolution thin-section photos, yielding results highly consistent with virtual microscopy.

\subsection{Convolutional Neural Networks (CNN)}

The architectural strength of CNNs lies in their ability to exploit spatial correlations through a hierarchical structure, utilizing convolutional layers for feature extraction and pooling layers for dimensional reduction \cite{goodfellow2016deep, lecun2015deep}. The evolution of these designs, including residual connections and dense blocks, has addressed the vanishing gradient problem and enabled substantial improvements in high-dimensional tasks \cite{zhang2022resnest}.

The versatility of CNNs is particularly evident in geoscientific data processing. Foundational developments in object detection \cite{ren2015faster, tan2020efficientdet} have paved the way for specialized implementations in the petroleum industry, such as lithology identification from borehole images \cite{zhang2017deep} and high-resolution hyperspectral mineral mapping \cite{Resolution2021}. These applications underscore the potential of deep learning to automate labor-intensive geological surveys and improve exploration precision. Unlike traditional techniques, CNNs integrate the learning of optimal representations directly into the training process, autonomously identifying relevant structural and textural features in geological imagery \cite{lecun2015deep}.

The versatility of these deep learning paradigms extends beyond geoscientific applications. Recent advances in adjacent high-resolution domains illustrate the breadth of GAN and CNN architectures: the frequency-to-spectrum mapping 
GAN~\cite{wang2023frequency} introduced effective semi-supervised strategies for anomaly detection, while global feature-injected blind-spot networks~\cite{wang2024global} and multi-granularity feature enhancement networks~\cite{ying2024multi} have improved hierarchical and spatial-spectral feature extraction in complex visual environments. These developments underscore the broader applicability of the architectural principles employed in our core analysis framework.

\subsection{Object Detection and Image Inpainting}

\subsubsection{Object Detection Frameworks}
Object detection has evolved significantly with the advent of deep learning. Innovative frameworks like Libra R-CNN \cite{Pang_2019_CVPR} address imbalances at the sample and feature levels using IoU-balanced sampling and balanced feature pyramids. Other architectures, such as ResNeSt \cite{zhang2022resnest}, incorporate channel-wise attention strategies (Split-Attention blocks) to capture complex interdependencies within feature maps. Furthermore, Dynamic R-CNN \cite{DynamicRCNN} adaptively adjusts the IoU threshold and regression loss shape during training, maximizing the quality of object proposals—a critical step in our pipeline for isolating damaged core sections.

\subsubsection{Image Inpainting and Texture Reconstruction}
Image inpainting is the process of reconstructing missing or damaged regions in a visually and semantically consistent manner. Early approaches included diffusion-based methods \cite{7987733} and exemplar-based techniques \cite{zhang2012exemplar}, which prioritized filling order based on local variances.

Modern deep generative models have transformed this field by synthesizing novel structures while referencing surrounding attributes. Gated convolutions \cite{yu2019free} provided a learnable dynamic feature selection mechanism, addressing the limitations of vanilla convolutions in handling irregular masks. For texture-heavy data like geological cores, orderless representations such as FV-CNN \cite{cimpoi2015deep} and bilinear models \cite{Lin_2015_ICCV} have proven effective in modeling local pairwise feature interactions. These advancements provide the foundation for the Contextual Residual Aggregation (CRA) technique used in our study to restore high-frequency textures in damaged core samples.

\section{Proposed Work}
\label{sec:prosedwork}

\begin{figure*}[t]
    \centering
    \includegraphics[width=\textwidth]{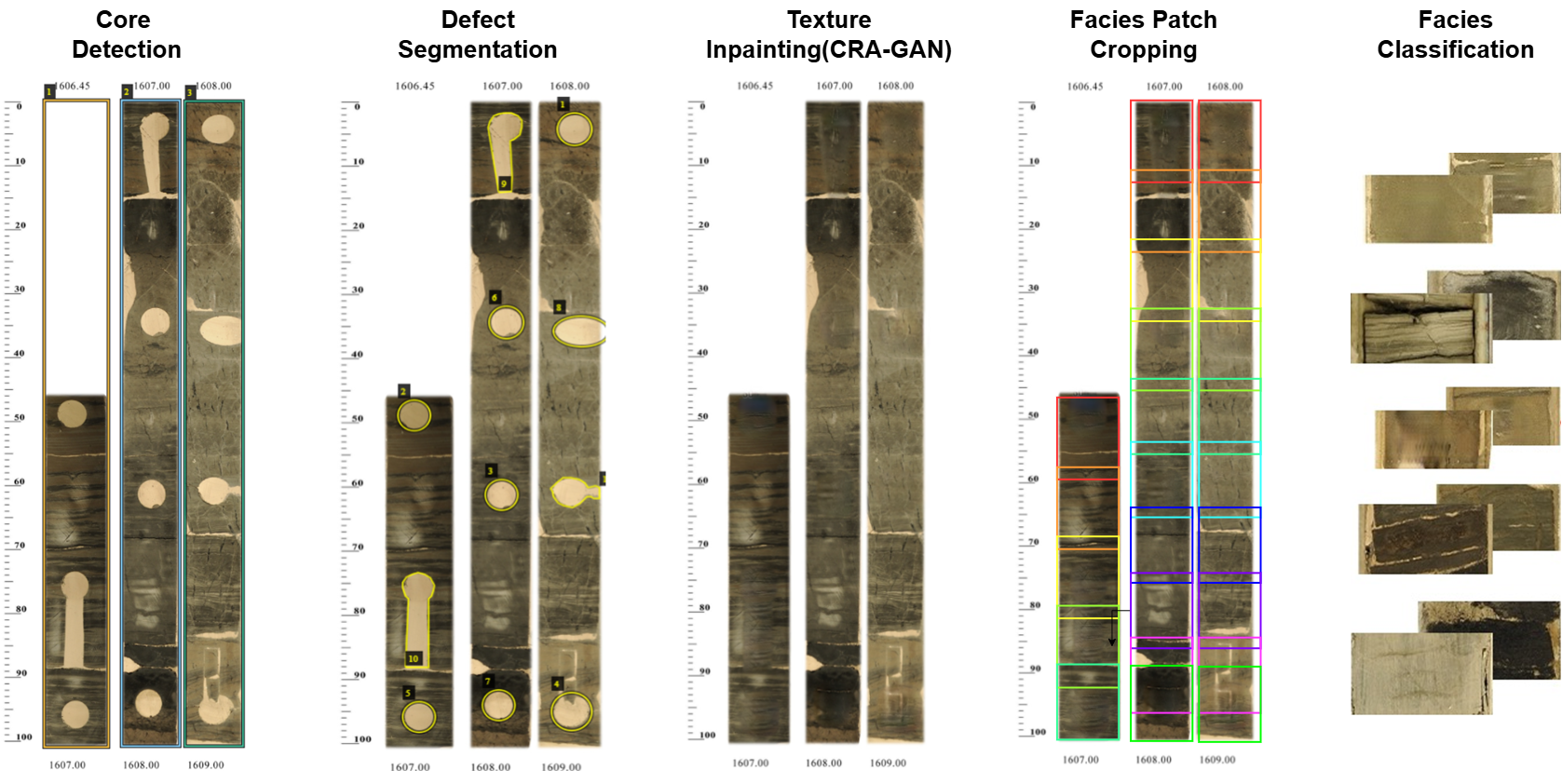} 
    \caption{High-level overview of the proposed geological core analysis framework, 
    illustrating the transition from raw input through automated detection and 
    inpainting to the final classification stage.}
    \label{fig:overall_pipeline}
\end{figure*}

The integrated framework for automated drill-core analysis is illustrated in 
Fig.~\ref{fig:overall_pipeline}. The pipeline consists of four sequential stages: 
(1) core column detection, (2) defect segmentation, (3) structural inpainting, and 
(4) lithological facies classification. By isolating damaged regions and restoring 
them via generative modeling, we aim to evaluate the feasibility of automated 
classification under conditions of incomplete physical information.

\subsection{Preprocessing: Detection and Segmentation}

To isolate the geological data from the background, we first employ a Faster 
R-CNN~\cite{ren2015faster} object detection model to localize the core columns. 
Subsequently, a Mask R-CNN~\cite{he2017mask} instance segmentation model is applied 
to identify physical defects, such as plug holes or fractures. This stage generates 
precise binary masks that define the target areas for reconstruction, ensuring that 
the inpainting process is restricted to damaged sections while the valid geological 
texture remains unaltered.

\subsection{Adversarial Reconstruction Framework}
\label{sec:adversarial}

The reconstruction engine is based on a GAN designed to synthesize high-frequency textures. We employ a PatchGAN discriminator~\cite{isola2017Image}, which, unlike global discriminators, penalizes structural inconsistencies at the scale of local image patches. This approach is 
essential for capturing the granular, stochastic nature of sandstone and clay textures.

As shown in the training loop in Fig.~\ref{fig:inpainting_model} (indicated by 
dashed lines), the adversarial process is governed by a multi-component objective 
function. The total loss $\mathcal{L}_{total}$ combines three terms\,---\,an
adversarial term, a pixel-wise reconstruction term, and an auto-encoding
self-reconstruction term\,---\,and is defined as:
\begin{equation}
    \mathcal{L}_{total} = \lambda_{adv}\,\mathcal{L}_{adv}(G, D) +
    \lambda_{L1}\,\mathcal{L}_{L1}(G) +
    \lambda_{AE}\,\mathcal{L}_{AE}(G)
\end{equation}
The discriminator $D$ plays a dual role: it receives the original "Ground Truth" 
images and the "Generated" outputs from the generator $G$ as inputs. By 
distinguishing between real and synthesized patches, it provides adversarial 
feedback that guides the generator toward producing statistically consistent 
textures. The adversarial loss $\mathcal{L}_{adv}$ is formulated as:
\begin{equation}
    \mathcal{L}_{adv} = \mathbb{E}_{x \sim p_{data}(x)}[\log D(x)] + 
    \mathbb{E}_{z \sim p_{z}(z)}[\log(1 - D(G(z)))]
\end{equation}
Simultaneously, the reconstruction term $\mathcal{L}_{L1} = \lVert G(z) - x \rVert_1$
enforces pixel-wise fidelity between the generated output and the ground-truth core
texture $x$, while the auto-encoding term
$\mathcal{L}_{AE} = \lVert G(\tilde{x}) - x \rVert_1$ regularizes the generator by
requiring it to faithfully reconstruct unmasked inputs $\tilde{x}$. This term
stabilizes training and prevents the contextual-aggregation pathway from drifting
away from the valid surrounding texture. The corresponding weights
$\lambda_{adv}=0.001$ and $\lambda_{L1}=\lambda_{AE}=1.2$
(Table~\ref{tab:hyperparameters}) balance these three objectives. This configuration
uses adversarial feedback only during training to improve texture fidelity while
maintaining reconstruction stability.

\subsection{Image Inpainting with Contextual Residual Aggregation (CRA)}
\label{sec:cra}
\begin{figure*}[!t]
    \centering
    \includegraphics[width=\textwidth,height=1\textheight,keepaspectratio]
    {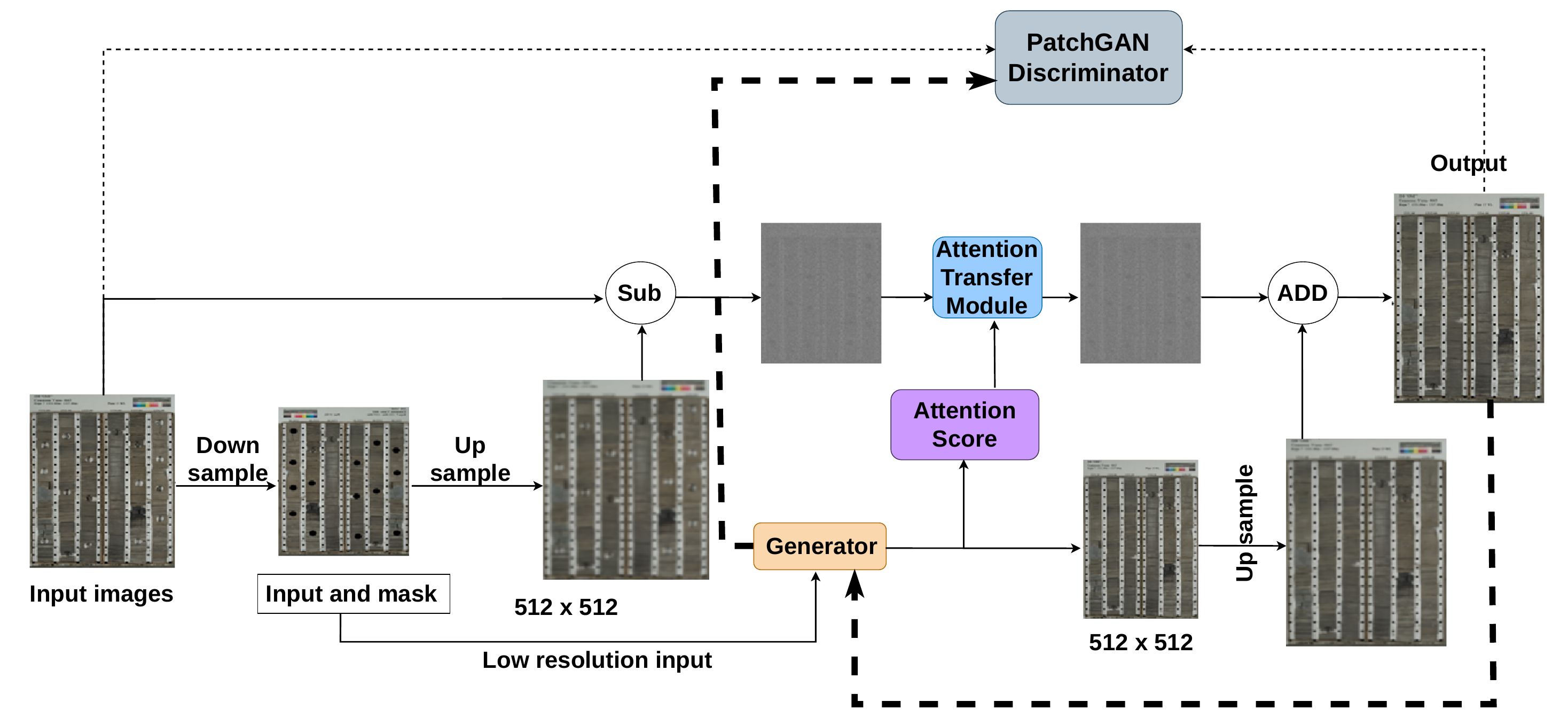}
    \caption{The proposed CRA-GAN inpainting pipeline. The architecture features a 
    dual-path generator guided by an Attention Score and a PatchGAN Discriminator. 
    The adversarial feedback loop (indicated by dashed lines) is active only during 
    training to improve texture fidelity. During inference, the discriminator is 
    omitted to reduce computational overhead.}
    \label{fig:inpainting_model}
\end{figure*}

To handle high-resolution imagery, we employ the Contextual Residual Aggregation 
(CRA) mechanism. As illustrated in Fig.~\ref{fig:inpainting_model}, the architecture 
splits the task into two complementary pathways: low-resolution content generation, 
which captures the global structural context of the core sample, and high-resolution 
texture refinement, which restores the grain-scale detail required for lithological 
discrimination. The internal Attention Computing Module (ACM) calculates contextual 
weights that guide the transfer of high-frequency residuals from unmasked regions to 
the damaged zones, ensuring that the synthesized content is statistically consistent 
with the surrounding valid texture. The final reconstructed output is obtained by 
combining these aggregated residuals with the upsampled low-resolution features. 
This dual-path approach makes the generator resolution-independent, enabling 
sharp outputs across a range of core image resolutions without retraining. The PatchGAN discriminator (Section~\ref{sec:adversarial}) is omitted during inference; the CRA generator operates as a standalone module at test time, reducing computational overhead for field deployment.

\subsection{Texture-Aware Classification Backbones}

\begin{figure*}[!h]
    \centering
    \includegraphics[width=\textwidth,height=\textheight,keepaspectratio]{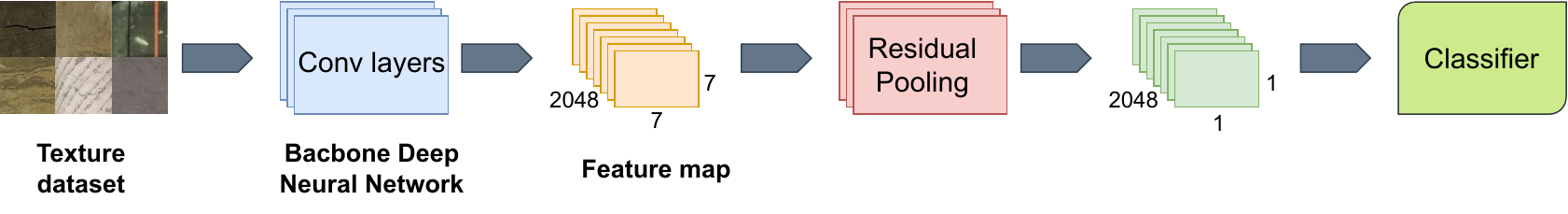}
    \caption{Architecture of the Deep Residual Pooling (DRP) model.}
    \label{fig:DRP-Texture_Recognition}
\end{figure*}

\begin{figure*}[!h]
    \centering
    \includegraphics[width=\textwidth,height=\textheight,keepaspectratio]{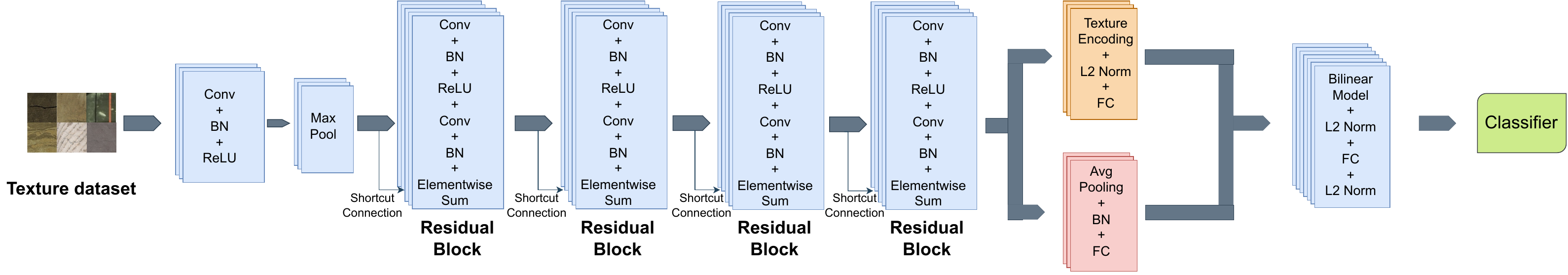}
    \caption{Architecture of the Deep Encoding Pooling Network (DEP) model.}
    \label{fig:DEP}
\end{figure*}

Once the images are restored, they are cropped into uniform patches for lithological 
classification. To evaluate the impact of reconstruction on classification, we select 
several specialized architectures designed for texture recognition:
\begin{enumerate}
    \item \textbf{Deep Residual Pooling (DRP):} We employ the DRP 
    framework~\cite{mao2021deep} (Fig.~\ref{fig:DRP-Texture_Recognition}), which 
    applies a simplified learnable residual encoding procedure. DRP treats features 
    from pre-trained CNN layers as a learned dictionary, generating orderless 
    representations via an aggregation module.
    \item \textbf{Deep Encoding Pooling (DEP):} To complement DRP, we also employ 
    the DEP network~\cite{xue2018deep} (Fig.~\ref{fig:DEP}). DEP integrates an 
    encoding layer for texture details with a global average pooling layer for 
    spatial context. By applying bilinear models to process these features, DEP 
    captures complex pairwise feature interactions.
\end{enumerate}
By benchmarking these specialized networks alongside modern Transformers (Swin, 
ViT), we assess how GAN-based reconstruction affects different feature extraction 
paradigms.

\section{Experimental Results}
\label{sec:approach-multi-modal}

\subsection{Dataset Preparation}

\begin{figure}[!h]
    \centering
    \includegraphics[width=.8\linewidth,height=.8\textheight,keepaspectratio]{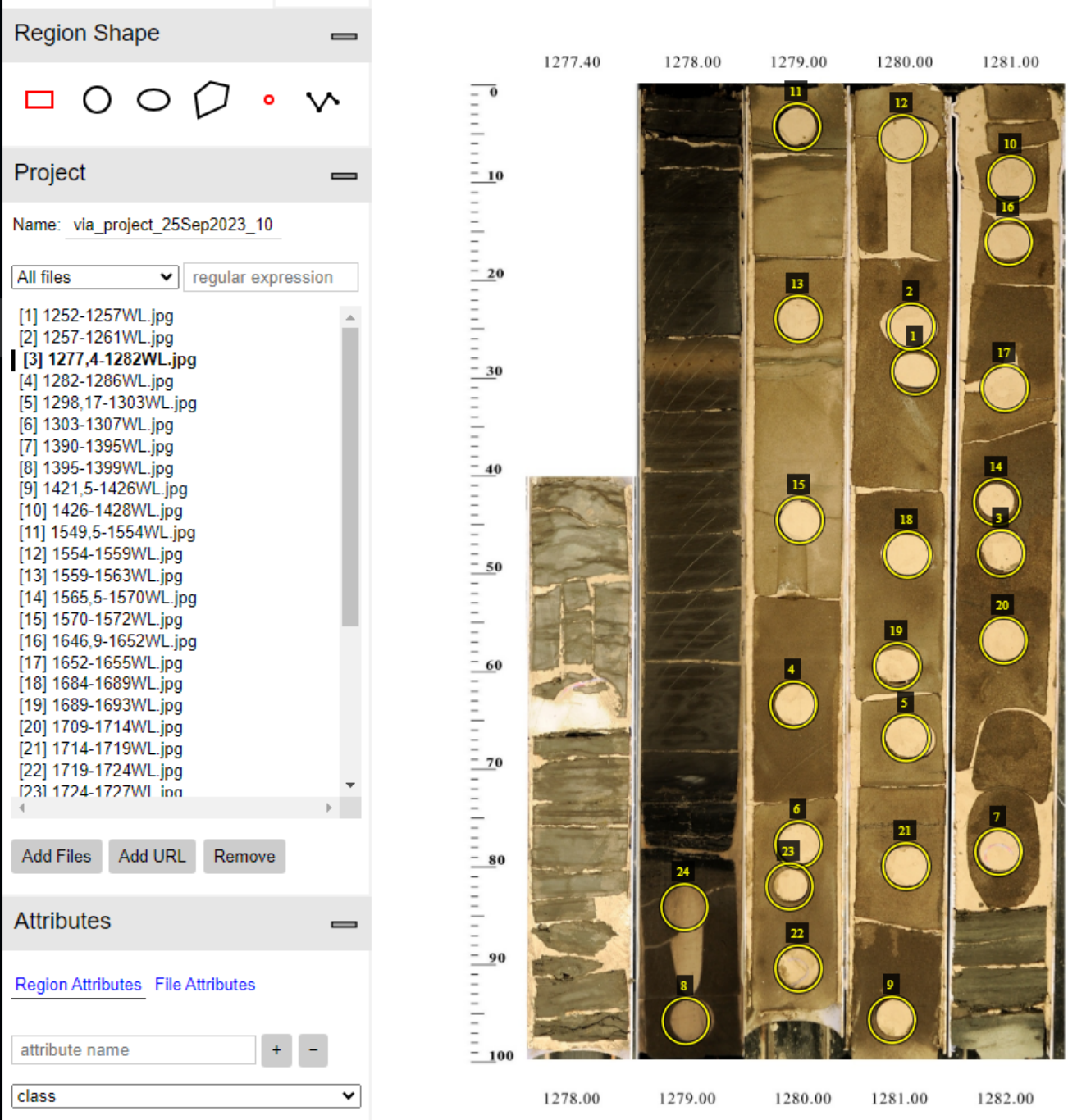}
    \caption{Tools for marking lithotypes on the image.}
    \label{fig:tools}
\end{figure}

The construction of a robust dataset was fundamental to evaluating the proposed 
framework. To facilitate efficient data labeling, we developed a customized 
web-based annotation toolkit based on the VGG Image Annotator 
(VIA)~\cite{dutta2019vgg}. By integrating the VIA core into a Flask-based 
environment, we established a centralized platform for multi-user processing 
and rapid iteration (see Fig.~\ref{fig:tools}).

\begin{figure}[!h]
    \centering
    \includegraphics[width=.8\linewidth,height=.8\textheight,keepaspectratio]{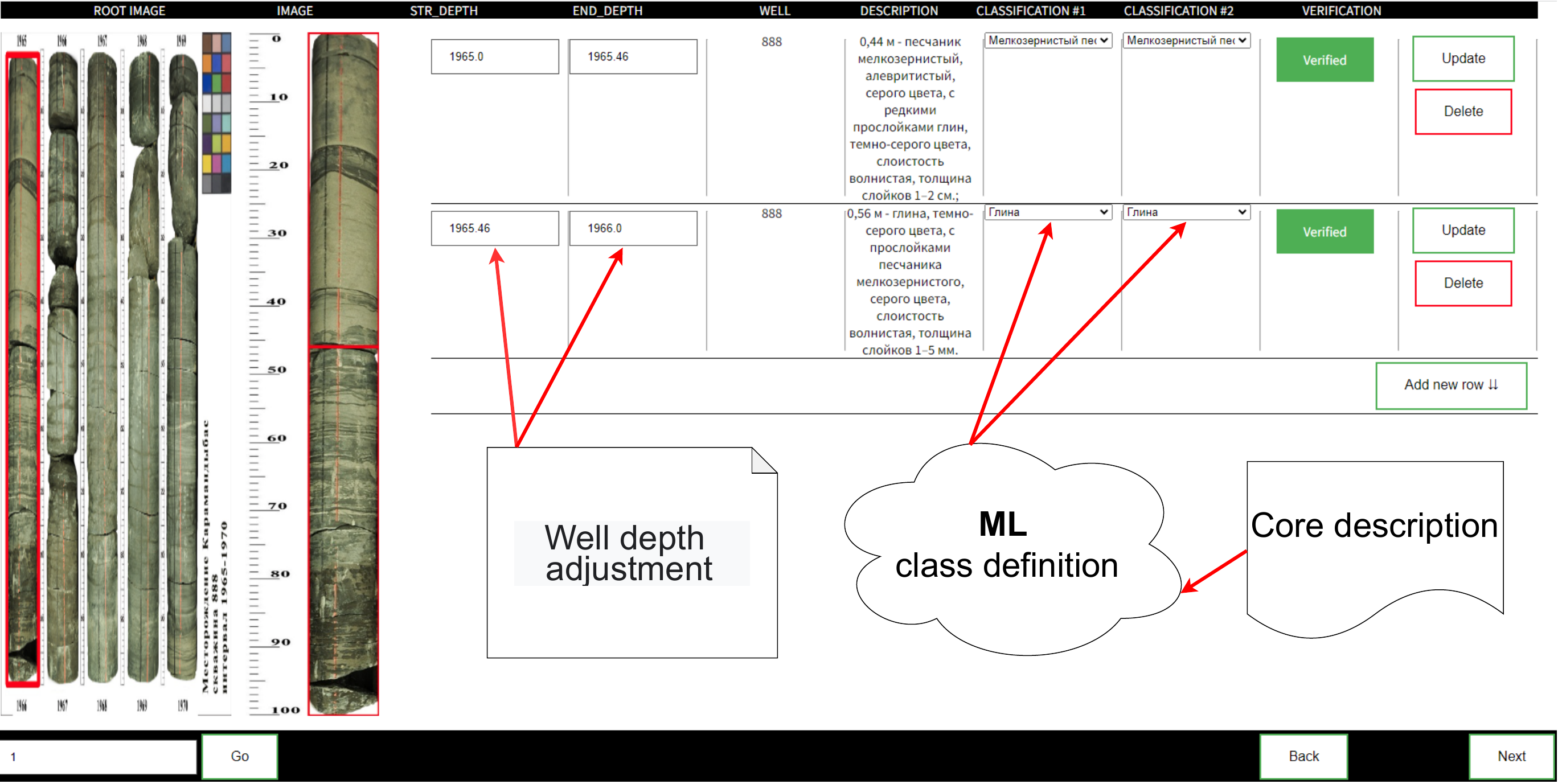}
    \caption{Dataset Image Markup Toolkit.}
    \label{fig:tools2}
\end{figure}

To accelerate the annotation process, we implemented a hybrid labeling strategy. 
An NLP-based model was first applied to extract preliminary lithology classes from 
text-based geological reports for an initial subset of five wells. These automated labels 
were then used to pre-fill attributes in the annotation tool 
(Fig.~\ref{fig:tools2}), after which domain experts performed manual verification 
and correction to ensure geological accuracy.

The resulting dataset comprised 2,190 high-resolution images of 1-meter drill-core 
sections obtained from a deposit in South-Western Kazakhstan. The dataset covered 
nine distinct lithological facies, providing a representative benchmark for texture 
analysis under realistic conditions of data loss and physical damage.

The complete dataset originated from sixteen geological wells within this single deposit. All 2{,}190 one-meter core images, and all experiments reported in this study, were derived from this same set of wells. The five wells referenced above denote only the subset to which NLP-assisted pre-labeling was applied, and do not constitute a separate dataset.

\subsection{Implementation Details}
\label{sec:implementation_details}

\subsubsection{Hardware and Software Environment}

The computational pipeline was implemented using a multi-framework strategy to 
leverage optimized baseline implementations. The CRA-GAN reconstruction module 
was developed in TensorFlow v1.13 to maintain compatibility with the original 
architecture's contextual aggregation layers; this legacy version was retained 
deliberately, as the CRA mechanism relies on low-level graph execution features 
unavailable in later releases. The primary classification backbones (ViT, Swin, 
EfficientNetV2, and ConvNeXt) were implemented in TensorFlow 
v2.14~\cite{abadi2016tensorflow}, while specialized texture recognition components 
— specifically DEP and DRP — were developed using the PyTorch 
framework~\cite{NEURIPS2019_9015} to leverage its native feature extraction 
capabilities.

All experiments were executed on a workstation equipped with dual Intel Xeon Gold 
5218R CPUs, an NVIDIA Quadro RTX 5000 GPU (16~GB VRAM), and 200~GB of RAM. This 
hardware configuration supported the efficient training and inference of 
high-resolution imagery ranging from 512 to 4K pixels.

\subsubsection{Training Protocols and Hyperparameters}

To ensure rigorous comparability and reproducibility, a consistent experimental 
environment was maintained across all computational stages. All classification 
models were optimized using the Adam algorithm. Learning rates were tuned 
independently for each architecture family to ensure stable convergence: 
$1\times10^{-4}$ for Transformer-based models (ViT and Swin) and $1\times10^{-3}$ 
for convolutional models (EfficientNetV2 and ConvNeXt). To mitigate overfitting, 
$L_2$ regularization with a weight decay of $1\times10^{-5}$ was applied to all 
classification heads.

Training duration was determined by architecture complexity. Swin and ViT reached 
convergence at 88 and 100 epochs, respectively, whereas EfficientNetV2 and ConvNeXt 
required 500 epochs to stabilize. Specialized texture models (DRP and DEP) were 
also trained for 500 epochs with a learning rate of 0.01.

For the CRA-GAN inpainting module, the input resolution was fixed at 
$512\times512$ pixels. Adversarial training stability was maintained by setting 
the discriminator momentum to $\beta_1 = 0.5$. The composite objective function 
employed loss coefficients of $\lambda_{adv} = 0.001$, $\lambda_{L1} = 1.2$, and 
$\lambda_{AE} = 1.2$ for the adversarial, reconstruction, and auto-encoding terms, 
respectively. A consolidated summary of all hyperparameters is provided in 
Table~\ref{tab:hyperparameters}.

\begin{table}[htbp]
\caption{Consolidated Hyperparameter Configurations for the Proposed Pipeline.}
\label{tab:hyperparameters}
\centering
\resizebox{\columnwidth}{!}{%
\begin{tabular}{lcc}
\hline
Hyperparameter & Classification Backbones & Inpainting Module \\ \hline
Architectures & Multiple Architectures & Generative Model \\
Framework & TensorFlow 2.14 / PyTorch & TensorFlow 1.13 \\
Input Resolution & $224 \times 224 \times 3$ & $512 \times 512 \times 3$ \\
Optimizer & Adam ($\beta_1=0.9$) & Adam ($\beta_1=0.5$) \\
Learning Rate & $10^{-4}$--$10^{-3}$ (Specialized: 0.01) & $1 \times 10^{-4}$ \\
Weight Decay ($L_2$) & $1 \times 10^{-5}$ & N/A \\
Batch Size & 32 / 64 (Specialized: 128) & 4 / 8 \\
Loss Coefficients & Cross-Entropy & $\lambda_{adv}=0.001, \lambda_{L1,AE}=1.2$ \\
Training Duration & 88, 100, 500 Epochs & 30,000 Steps \\
Augmentations & Resize, Flip, Color Jitter & Random Crop, Masking \\ \hline
\end{tabular}
}
\end{table}

\subsection{Evaluation Metrics and Statistical Validation}
\label{sec:metrics_analysis}

The performance evaluation was based on the following metrics.

\textbf{Overall Accuracy (Acc):}
\begin{equation}
Acc = \frac{\sum_{i=1}^C TP[i]}{\sum_{i=1}^C \sum_{j=1}^C CM[i][j]} \times 100\%
\end{equation}
where $C$ is the number of classes, $TP[i]$ represents the True Positives for 
class $i$ (diagonal elements of the confusion matrix), and $CM[i][j]$ is the 
element in row $i$, column $j$ of the confusion matrix.

\textbf{Mean Class Accuracy (MCA):}
\begin{equation}
MCA = \frac{1}{C}\sum_{i=1}^{C} \frac{TP_i}{N_i} \times 100\%
\end{equation}
where $N_i$ is the total number of samples belonging to class $i$.

To ensure statistical validity and account for test set variance, 95\% confidence 
intervals (CI) were computed via bootstrapping (1,000 iterations) for all evaluated 
architectures. The consolidated performance comparison across all evaluated configurations 
is provided in Table~\ref{tab:model_comparison}.

\subsection{Core Detection and Mask Generation}
\label{sec:detection_masking}

To ensure high-quality reconstruction, we implemented an automated preprocessing 
stage to localize and mask damaged regions (e.g., plug holes) in the core imagery. 
This process utilized a specialized dataset of 317 core images with approximately 
3,000 manually annotated plug holes. To ensure high geometric precision, complex 
polygonal masks were employed to capture the exact physical boundaries of the 
data loss areas.

For automated mask generation, Faster R-CNN~\cite{ren2015faster} was integrated 
for core localization and Mask R-CNN~\cite{he2017mask} for instance segmentation. 
In the proposed pipeline, these architectures served as functional, auxiliary 
modules rather than primary methodological contributions. This distinction allowed 
the workflow to benefit from robust, state-of-the-art detection capabilities while 
maintaining the study's focus on the subsequent inpainting and lithological 
classification tasks.

\begin{figure}[!h]
    \centering
    \includegraphics[width=.8\linewidth,height=.8\textheight,keepaspectratio]{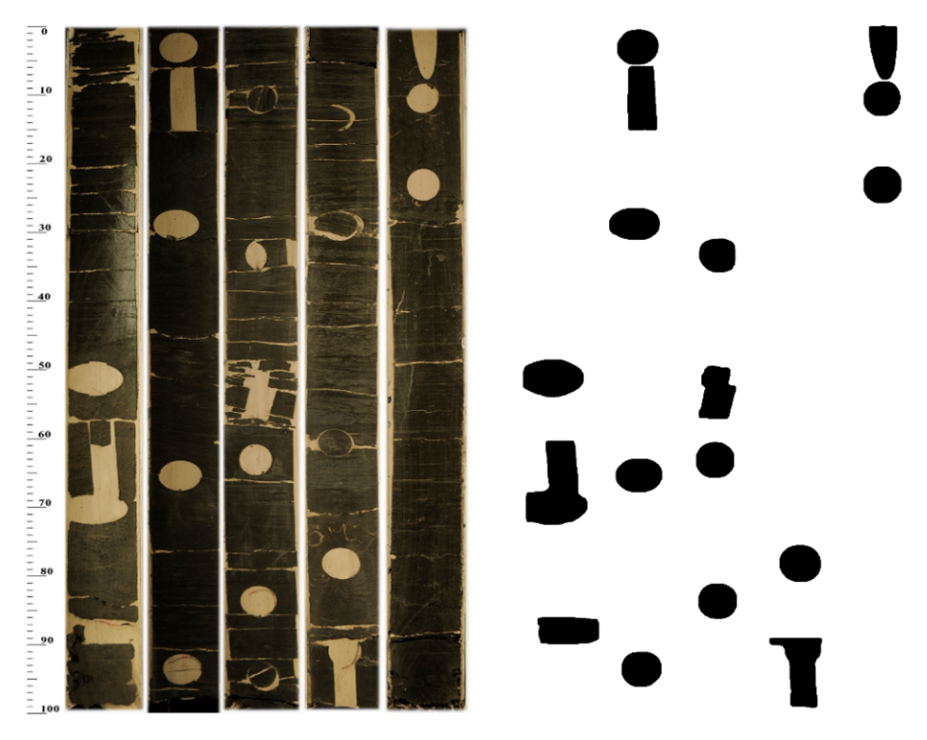}
    \caption{Automated detection of plug-sampling voids on field core 
    imagery, with identified defect regions highlighted by 
    instance segmentation masks generated via Mask R-CNN.}
    \label{fig:core_detection}
\end{figure}

Evaluation of these preprocessing modules demonstrated a detection accuracy of 
93\% for Faster R-CNN and a pixel-wise segmentation accuracy of 75\% for Mask 
R-CNN. These results provided sufficiently precise spatial constraints for the 
subsequent CRA-GAN phase. The integrated system automatically processed raw, physically damaged images to produce the binary masks required for reconstruction, as illustrated in Fig.~\ref{fig:core_detection}. 
This stage established a consistent digital foundation for the end-to-end 
lithological analysis.

\subsection{Hole Inpainting Implementation}
\label{sec:inpainting_details}

The CRA-GAN framework described in Section~\ref{sec:cra} was trained on 
core images with resolutions in multiples of 512 pixels. During processing, each 
input image was downsampled to $512\times512$ pixels to provide the global 
structural base for the dual-path generator. The contextual residuals computed 
by the ACM were added to the upsampled base, producing the final reconstructed 
output.

\subsection{Inpainting Quality Assessment}
\label{sec:inpainting_quality}

\begin{figure}[htbp]
    \centering
    \includegraphics[width=\linewidth]{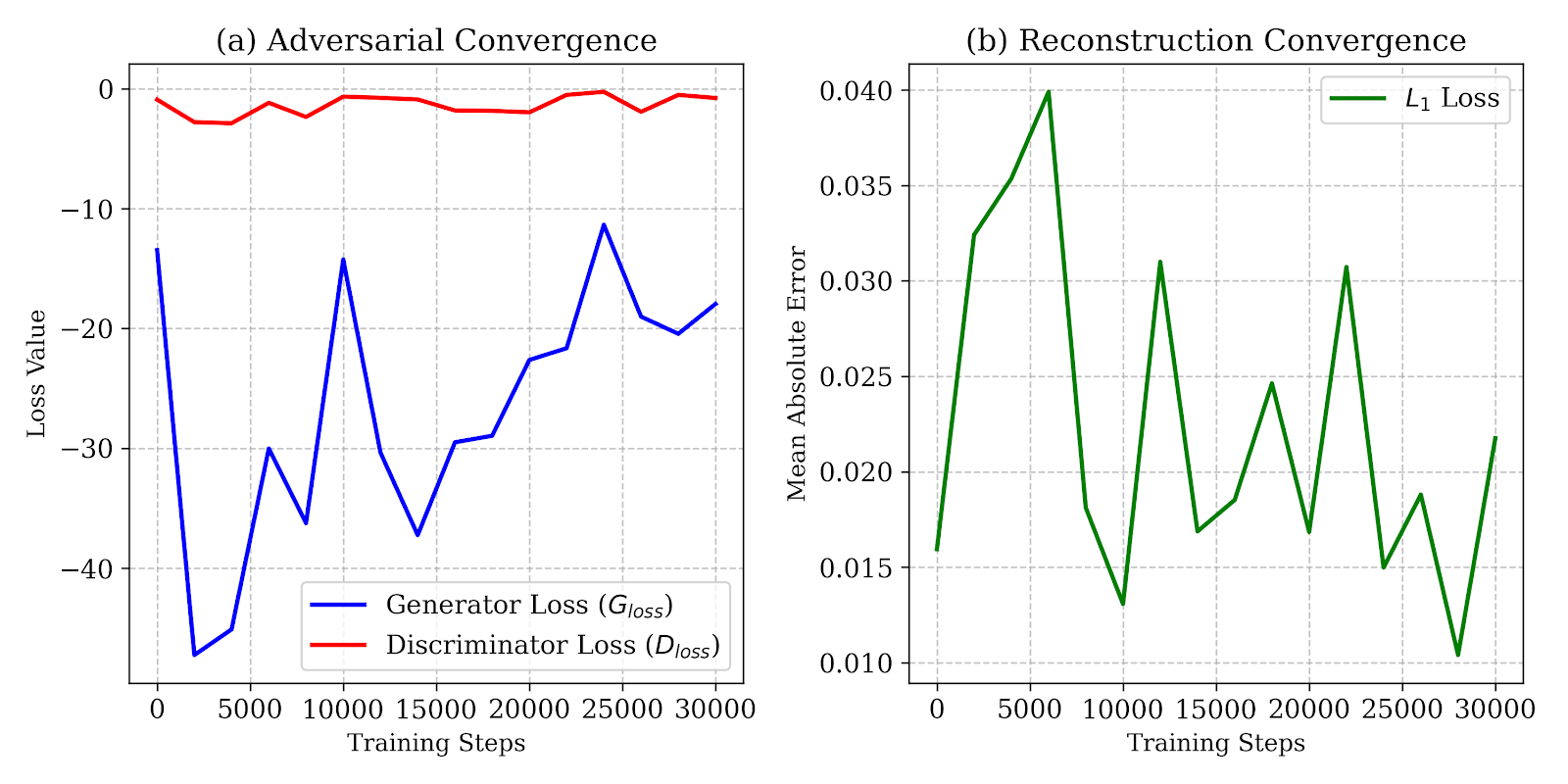}
    \caption{Experimental training dynamics of the CRA-GAN model: (a) Adversarial loss curves for the Generator ($G_{loss}$) and the Discriminator with gradient penalty ($D_{loss\_with\_gp}$); (b) L1 reconstruction loss trajectory over 30,000 training steps.}
    \label{fig:convergence}
\end{figure}

Training stability was verified through convergence analysis 
(Fig.~\ref{fig:convergence}). The adversarial process involving the PatchGAN 
discriminator~\cite{isola2017Image} exhibited stable dynamics over 30,000 
iterations, and the $L_1$ reconstruction loss declined to a plateau of 
approximately 0.02, indicating successful learning of the underlying texture 
distribution without mode collapse.

\begin{figure*}[t!] 
    \centering
    \begin{subfigure}{0.48\textwidth} 
        \centering
        \includegraphics[width=\linewidth]{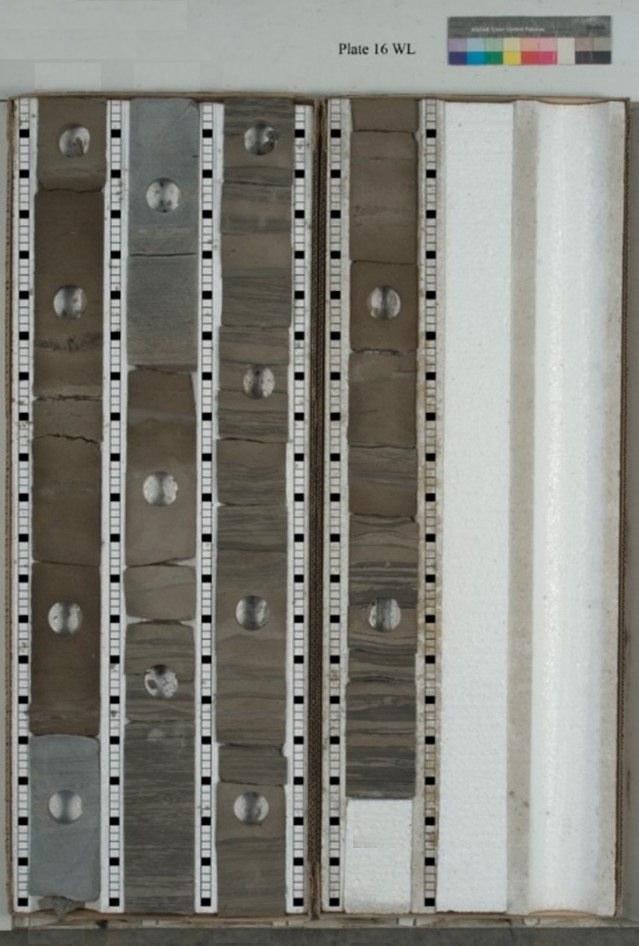}
        \caption{Input Image.}
        \label{fig:inp_1}
    \end{subfigure}
    \hfill 
    \begin{subfigure}{0.48\textwidth}
        \centering
        \includegraphics[width=\linewidth]{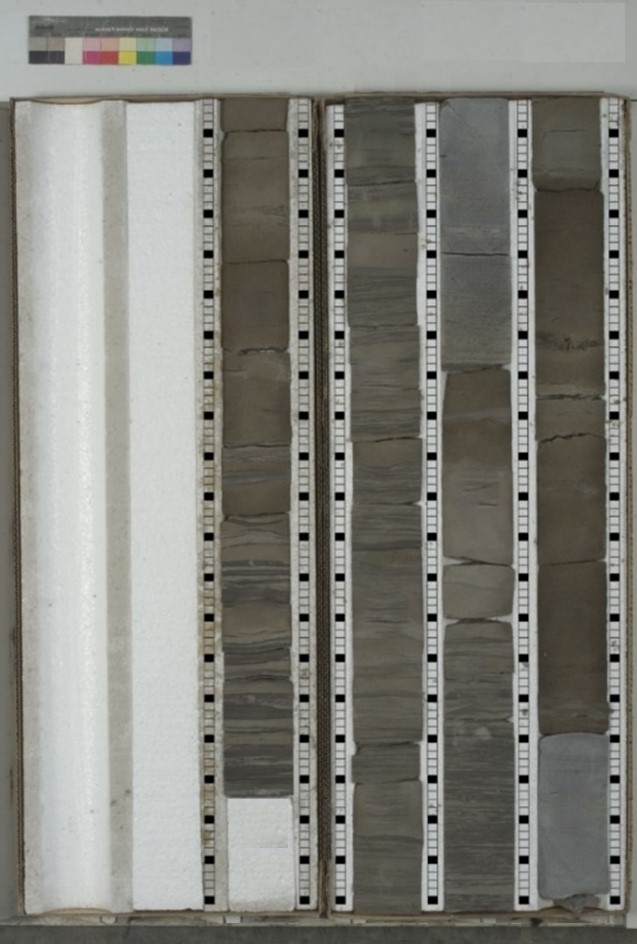}
        \caption{Output Image.}
        \label{fig:inp_2}
    \end{subfigure}
    
    \caption{Example of the Model output: (a) input image; (b) output image.}
    \label{fig:examples_images_1}
\end{figure*}

\begin{figure}[h!]
    \centering
    \begin{subfigure}{0.48\columnwidth}
        \includegraphics[width=\textwidth]{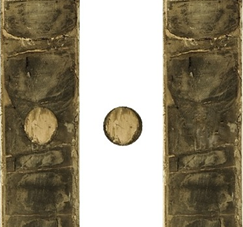}
        \caption{}
    \end{subfigure}\hfill
    \begin{subfigure}{0.48\columnwidth}
        \includegraphics[width=\textwidth]{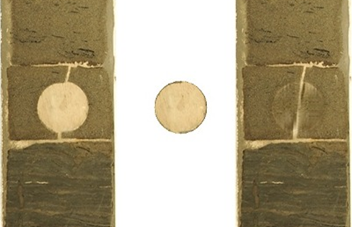}
        \caption{}
    \end{subfigure}\\[2pt]
    \begin{subfigure}{0.48\columnwidth}
        \includegraphics[width=\textwidth]{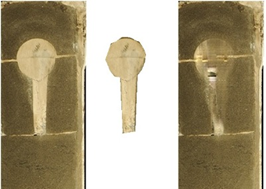}
        \caption{}
    \end{subfigure}\hfill
    \begin{subfigure}{0.48\columnwidth}
        \includegraphics[width=\textwidth]{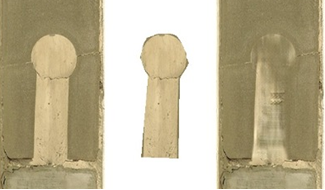}
        \caption{}
    \end{subfigure}
    \caption{Representative examples of CRA-GAN inpainting results: 
    left column --- original images with masked regions; 
    center column --- binary masks; right column --- 
    reconstructed outputs produced by the generator.
    } 
\label{fig:examples_images_2}
\end{figure}

Three complementary metrics were used for quantitative evaluation: Peak 
Signal-to-Noise Ratio (PSNR), Structural Similarity Index (SSIM), and 
Fr\'{e}chet Inception Distance (FID). The CRA-GAN achieved an average PSNR of 
28.7~dB and an SSIM of 0.91, indicating high pixel-level and structural fidelity. 
The FID score of 74.01 confirmed that the reconstructed textures were perceptually 
plausible (Fig.~\ref{fig:examples_images_1} and Fig.~\ref{fig:examples_images_2}), 
though a residual distributional distance remained. The implications of this gap 
for downstream classification are examined in Section~\ref{sec:discussion}.

\subsection{Comparative Analysis of Inpainting Architectures}
\label{sec:baseline_analysis}

A critical concern in geological computer vision pipelines was whether the inpainting 
stage introduced "hallucinations" that could bias the classification backbone. To 
justify the selection of the CRA-GAN architecture, a comparative study was conducted 
against three established models: AOT-GAN, ICT, and Pluralistic Inpainting. This 
analysis also clarified the role of the inpainting stage relative to a 
"no-inpainting" baseline, in which damaged images were classified without prior 
reconstruction.

\subsubsection{Theoretical Limitations of Masked Inputs}

Bypassing the inpainting stage by classifying images with black or mean-filled masks 
introduced artificial high-frequency edges at the boundaries of the voids. For 
attention-based backbones such as the Swin Transformer, these non-geological 
discontinuities acted as structural noise, biasing the attention maps and preventing 
the network from focusing on the underlying rock texture. While the ablation study 
presented in Section~\ref{sec:ablation_analysis} demonstrated that global 
classification accuracy converged to 53\% regardless of whether inpainting was 
applied, this aggregate metric obscured class-level effects of practical geological 
significance. Specifically, CRA-GAN reconstruction yielded a 3.5-fold improvement 
in recall for the \textit{Dense rock} class (F1: 0.03~$\rightarrow$~0.11) and a 
notable gain for \textit{Medium-grained sandstone} (F1: $+$0.05), confirming that 
structural continuity restoration was beneficial for lithofacies whose discriminative 
features depended critically on grain-scale texture. Global accuracy was therefore limited primarily by inter-class textural overlap rather than by data completeness. The role of CRA-GAN as a preprocessing stage is quantified in the ablation study (Section~\ref{sec:ablation_analysis}).

\subsubsection{Failure Modes of Baseline Models}

\begin{figure*}[t!]
    \centering
    \begin{subfigure}{0.445\textwidth}
        \centering
        \includegraphics[width=\linewidth]{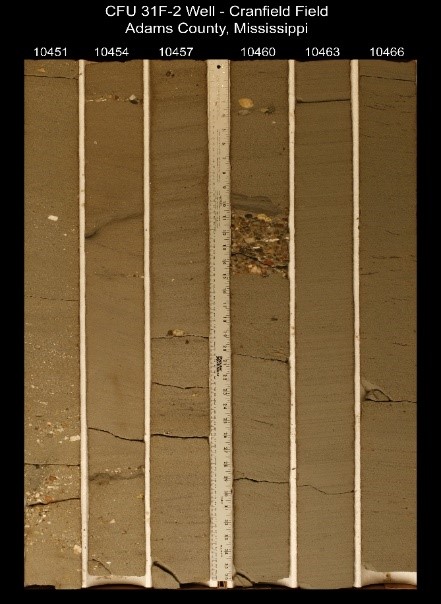}
        \caption{Input Image.}
    \end{subfigure}
    \hfill 
    \begin{subfigure}{0.48\textwidth}
        \centering
        \includegraphics[width=\linewidth]{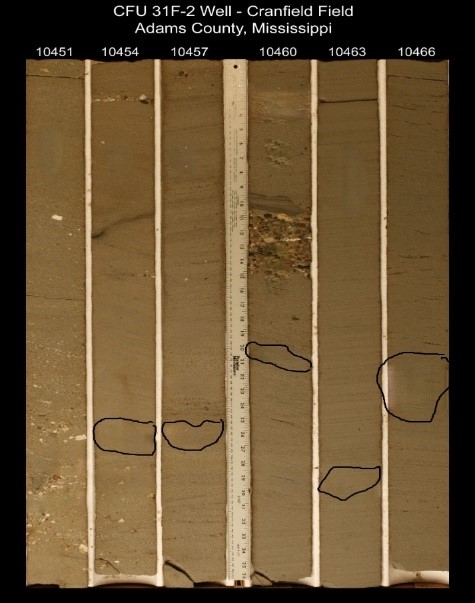}
        \caption{Output Image.}
    \end{subfigure}
    
    \caption{AOT-GAN results: (a) input image; (b) output image demonstrating structural blurring in high-resolution regions.}
    \label{fig:failure_aot}
\end{figure*}

\begin{figure}[htbp]
    \centering
    \includegraphics[width=1.0\linewidth, height=0.48\textheight]{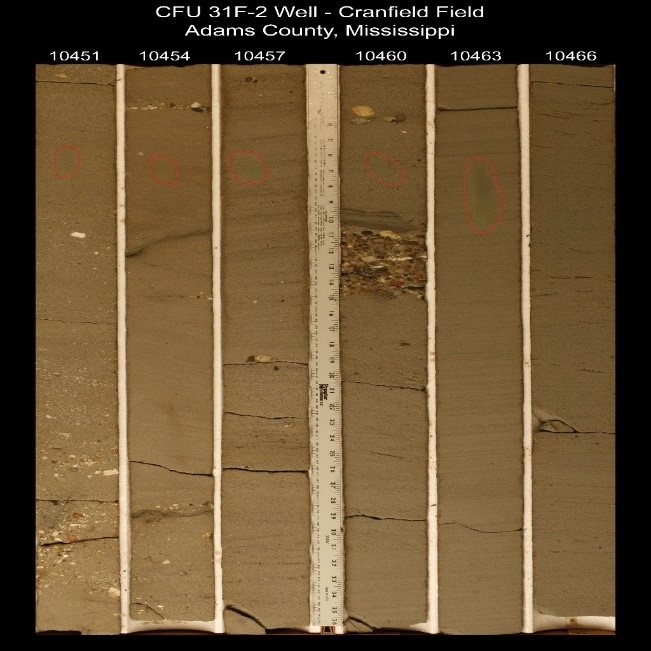}
    \caption{ICT (Transformer) results: Even at a reduced scale, the reconstruction exhibits a lack of high-frequency details, producing an overly smoothed texture that obscures lithological features.}
    \label{fig:failure_ict}
\end{figure}

\begin{figure}[htbp]
    \centering
    \begin{subfigure}{0.48\linewidth}
        \centering
        \includegraphics[width=\textwidth]{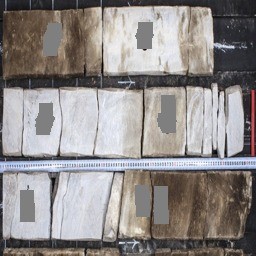}
        \caption{} 
    \end{subfigure}
    \hfill
    \begin{subfigure}{0.48\linewidth}
        \centering
        \includegraphics[width=\textwidth]{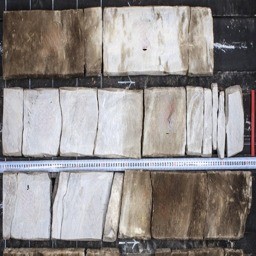}
        \caption{} 
    \end{subfigure}
    \caption{Pluralistic GAN results: (a) and (b) show severe chromatic artifacts (reddish tints) that could be misinterpreted as mineral oxidation.}
    \label{fig:failure_plural}
\end{figure}

The comparative evaluation revealed several systematic failures in baseline 
architectures:

\begin{itemize}
    \item \textbf{Structural Blurring and Scale Sensitivity (AOT-GAN):} The AOT-GAN 
    model exhibited significant performance degradation when applied to 
    high-resolution imagery ($>2000 \times 2000$ pixels). As illustrated in 
    Fig.~\ref{fig:failure_aot}, the transition from input 
    (Fig.~\ref{fig:failure_aot}a) to output (Fig.~\ref{fig:failure_aot}b) was 
    marked by severe textural blurring. This loss of grain-level stochastic patterns 
    prevented the classifier from detecting fine-grained facies, often leading to 
    the misclassification of sandstones as clay-rich structures.

    \item \textbf{Oversmoothing in Transformer-based Inpainting (ICT):} Results 
    from the Image Completion Transformer (ICT) highlighted the limitations of 
    self-attention in preserving local variance for textures. As shown in 
    Fig.~\ref{fig:failure_ict}, the model produced an overly smoothed appearance. 
    By acting as a low-pass filter, the ICT suppressed the diagnostic 
    "roughness" of the facies, which constituted a critical feature for 
    lithological discrimination.

    \item \textbf{Chromatic Artifacts and Mineralogical Bias (Pluralistic 
    Inpainting):} The most significant failure for industrial applications was 
    observed in the Pluralistic GAN. As evidenced in Fig.~\ref{fig:failure_plural}, 
    the model introduced severe chromatic aberrations, specifically unnatural 
    reddish tints. In geological logging, color is a primary indicator of 
    mineralogical composition; thus, such "hallucinated" chromatic shifts could 
    be misinterpreted as mineral oxidation (e.g., hematite), adversely affecting 
    downstream geological interpretation.
\end{itemize}

The inability of baseline models to maintain high-frequency texture and chromatic 
integrity underscored the necessity of the proposed CRA-GAN framework. In contrast 
to these alternatives, CRA-GAN ensured that reconstructed regions remained 
geologically consistent, thereby preserving the class-discriminative features 
required for reliable lithofacies classification — particularly for structurally 
complex minority classes where the impact of reconstruction quality was most 
pronounced.

\subsection{Classification Experimental Design and Results}
\label{sec:classification_results}

\begin{figure}[!h]
    \centering
    \includegraphics[width=0.6\linewidth,height=\textheight,keepaspectratio]{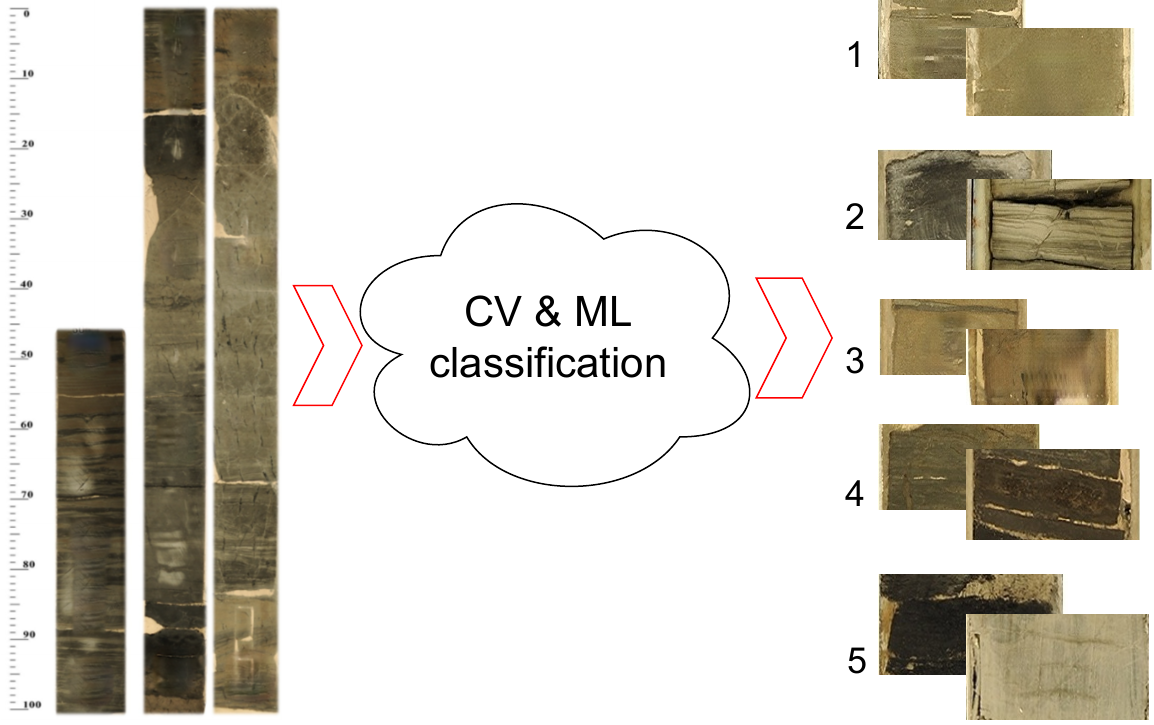}
    \caption{Representative patches of the distinct facies prepared for the model training phase.}
    \label{fig:exp_1}
\end{figure}

The classification of heterogeneous lithofacies was conducted in multiple stages, 
reflecting an evolution from basic categorical identification to granular facies 
analysis, coupled with a progressive shift toward more rigorous validation 
protocols. To ensure a fair and consistent evaluation across all architectures, 
the dataset preparation process was standardized: raw 1-meter core images were 
cropped into patches of $100 \times 100$ pixels using a sliding window stride of 
20 pixels (Fig.~\ref{fig:exp_1}). The extracted patches were randomly partitioned into training (70\%), 
validation (15\%), and testing (15\%) sets unless explicitly stated otherwise. The 
class distribution reflected natural geological occurrences, resulting in an 
inherently imbalanced dataset in which clay and sandstone dominated, while rare 
facies such as coal and dense rock were underrepresented.

\subsubsection{Initial 6-Class Baseline and DRP Performance}

\begin{figure}[!h]
    \centering
    \includegraphics[width=0.8\linewidth,height=\textheight,keepaspectratio]{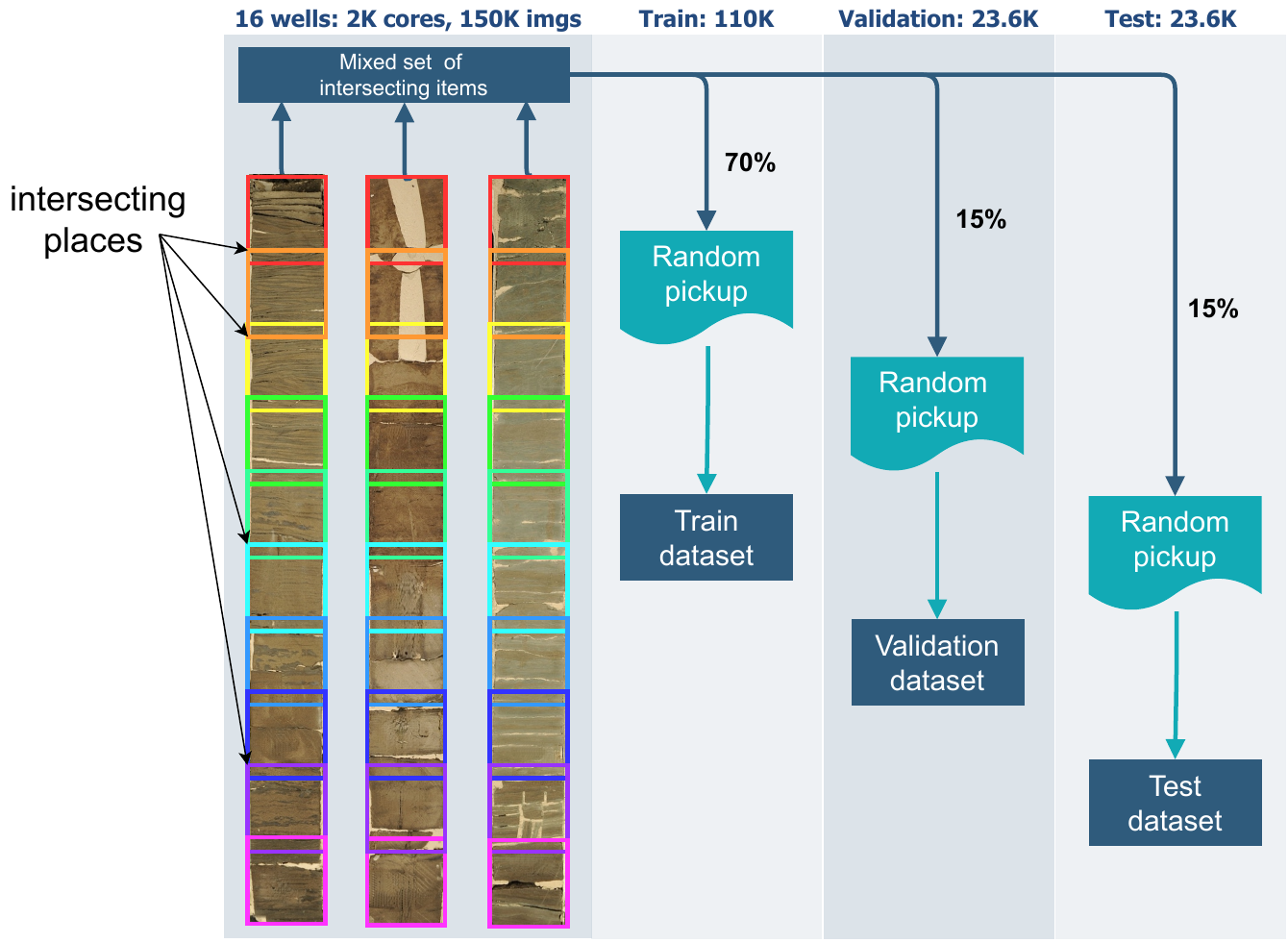}
    \caption{Dataset partition demonstrating the randomized split approach.}
    \label{fig:cores_ds_2}
\end{figure}

\begin{figure}[!h]
    \centering
    \includegraphics[width=\linewidth,height=\textheight,keepaspectratio]{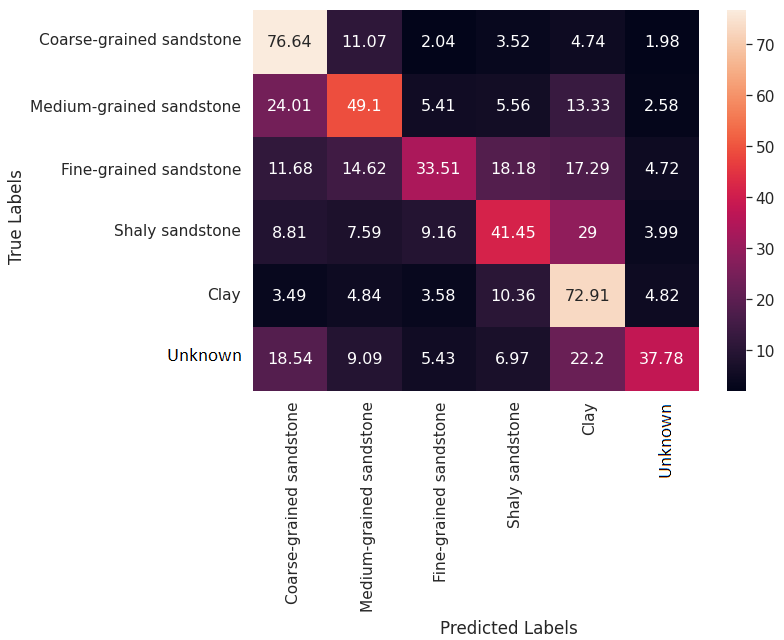}
    \caption{Deep Residual Pooling network confusion matrix for the initial 5-class (plus one indeterminate) dataset.}
    \label{fig:cm_drp_5classes}
\end{figure}

In the preliminary phase, the classification task was focused on six categories: 
five primary facies types and one indeterminate class. To establish a baseline for 
texture recognition, the Deep Residual Pooling (DRP) network~\cite{mao2021deep} 
with a ResNet-50 backbone was employed. Under a standard random partitioning 
strategy (70\% training, 15\% validation, and 15\% testing), as shown in 
Fig.~\ref{fig:cores_ds_2}, the DRP network achieved an overall accuracy of 60.0\% 
with a Mean Class Accuracy (MCA) of 51.9\% (Fig.~\ref{fig:cm_drp_5classes}).

Two distinct partitioning regimes are employed in this study and must not be
conflated. The first is a random patch-level split (70\%/15\%/15\% for training,
validation, and testing), retained solely as a diagnostic to expose spatial data
leakage. The metrics obtained under this regime are reported only to quantify the
inflation artifact, and are not used to support any final performance claim. The
second is a strict cross-well split, in which all patches from a given well are
assigned exclusively to a single partition, such that the two held-out test wells
share no spatial overlap with the fourteen training wells. All headline metrics in
Table~\ref{tab:model_comparison}, and all per-class results in
Sections~\ref{sec:arch_benchmark}--\ref{sec:ensemble}, are derived exclusively from
this single fixed cross-well split. The split was generated once and held constant
across every architecture to ensure comparability. The 95\% confidence intervals
(1{,}000 bootstrap iterations) therefore quantify test-set sampling variance under
this fixed split, rather than variance across repeated splits.

\subsubsection{Expansion to 9 Classes and the Data Leakage Phenomenon}

\begin{figure}[!h]
    \centering
    \includegraphics[width=\linewidth,height=\textheight,keepaspectratio]{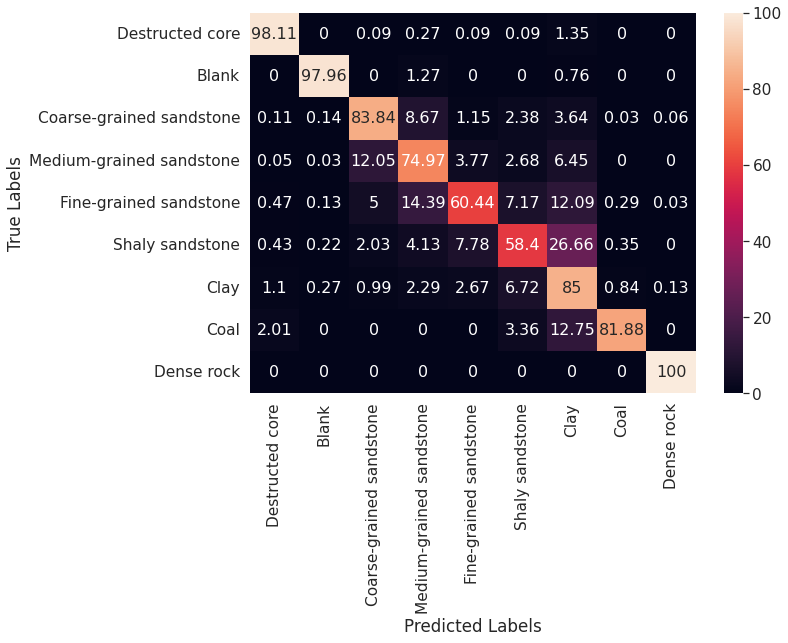} 
    \caption{Deep Residual Pooling network confusion matrix for the expanded 9-class dataset. This is a random-split result, inflated by data leakage and shown for diagnostic purposes only.}
    \label{fig:cm_drp_9classes}
\end{figure}

\begin{figure}[!h]
    \centering
    \includegraphics[width=\linewidth,height=\textheight,keepaspectratio]{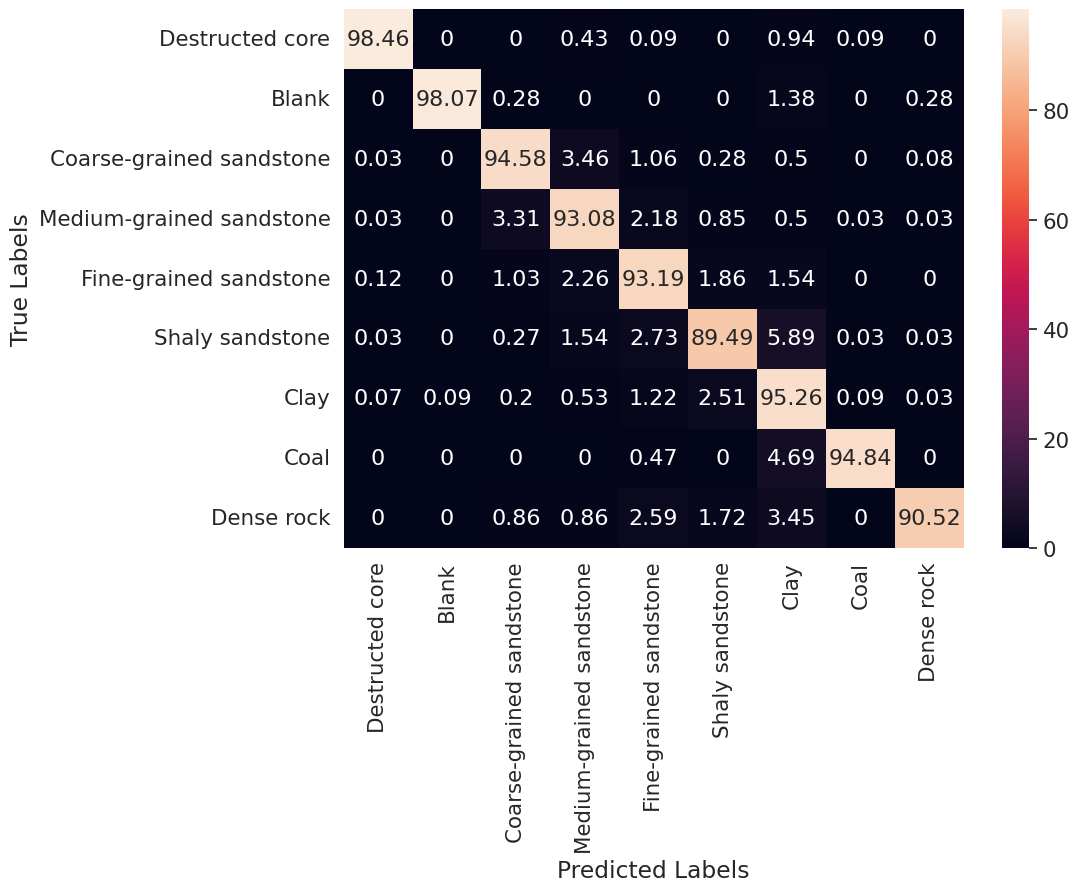}
    \caption{Confusion matrix of the DEP model on the randomly split dataset. The inflated accuracy (93.7\%) highlights the effect of data leakage prior to applying cross-well validation. Reported for diagnostic purposes only and excluded from the final cross-well comparison.}
    \label{fig:cm_3}
\end{figure}

To capture finer geological nuances, the taxonomy was expanded to nine highly 
granular lithological categories: destructed core, sandstone variants 
(coarse/medium/fine-grained and shaly), clay, coal, and dense rock. Using this 
expanded dataset under the same random split strategy, the DRP model reached an 
apparent accuracy of 75.8\% (Fig.~\ref{fig:cm_drp_9classes}), while the Deep 
Encoding Pooling (DEP) architecture~\cite{xue2018deep} achieved an extraordinarily 
high accuracy of 93.7\% (Fig.~\ref{fig:cm_3}).

However, a critical post-hoc analysis revealed that these results were artificially
inflated by spatial data leakage. Accordingly, they are reported here strictly as a
diagnostic, and are excluded from the consolidated cross-well comparison in
Table~\ref{tab:model_comparison}. Because the core patches were extracted using a 
sliding window with high overlap (100-pixel patch height with a 20-pixel stride), 
the random split allowed visually near-identical features from the same core 
sections to appear in both the training and testing sets, leading to severe 
overfitting. This finding necessitated a transition to a more rigorous validation 
strategy.

\subsubsection{Rigorous Cross-Well Validation and Honest Baseline}

\begin{figure}[!h]
    \centering
    \includegraphics[width=0.8\linewidth,height=\textheight,keepaspectratio]{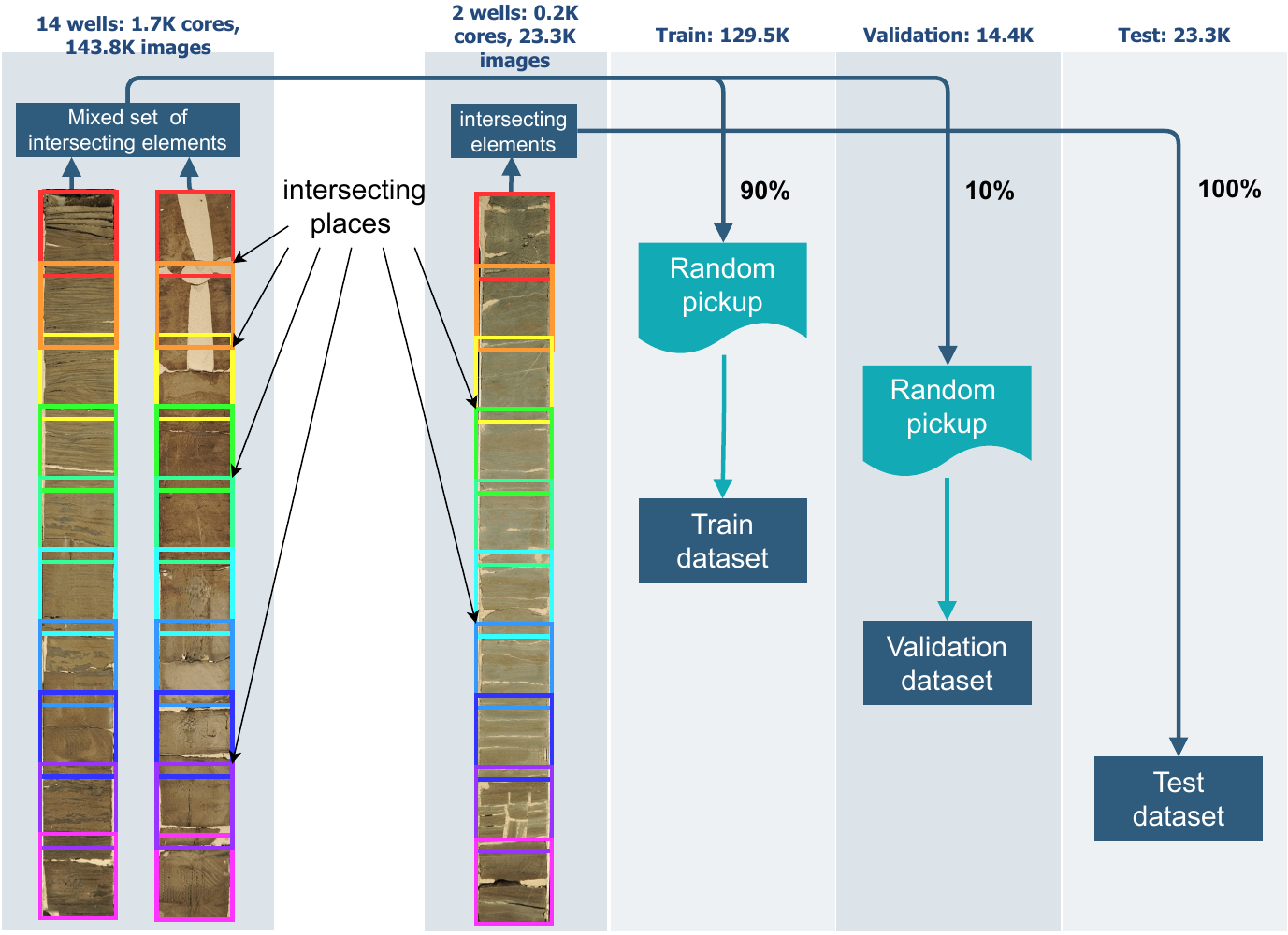}
    \caption{Strict cross-well validation split, ensuring that test images are extracted from entirely distinct geological wells unrepresented in the training data.}
    \label{fig:cores_ds_3}
\end{figure}

\begin{figure}[!h]
    \centering
    \includegraphics[width=\linewidth,height=\textheight,keepaspectratio]{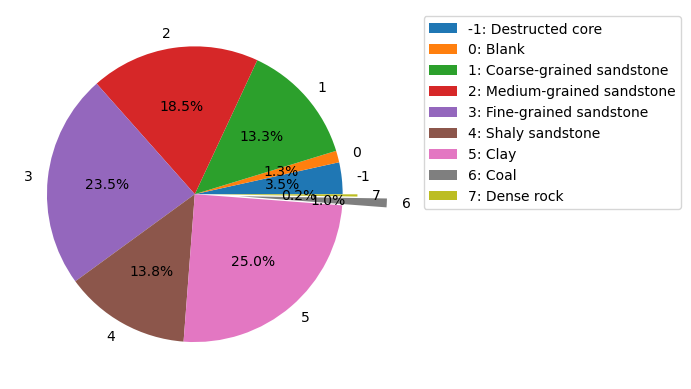}
    \caption{Class distribution of the strictly separated training dataset, demonstrating the heavy dominance of clay and sandstone facies.}
    \label{fig:distrib_1}
\end{figure}

\begin{figure}[!h]
    \centering
    \includegraphics[width=\linewidth,height=\textheight,keepaspectratio]{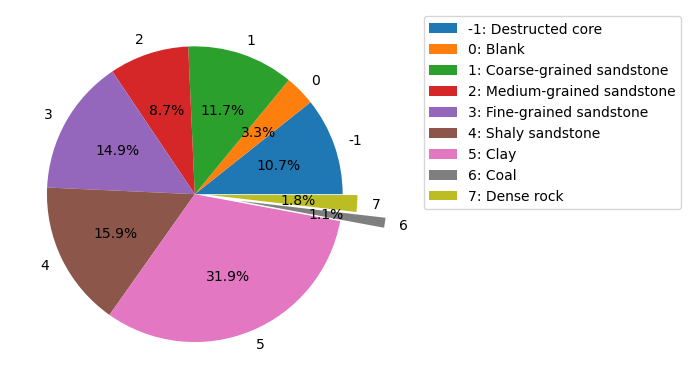}
    \caption{Class distribution of the isolated cross-well test dataset.}
    \label{fig:distrib_2}
\end{figure}

To address the data leakage issue and obtain a realistic assessment of model 
generalization, the methodology was overhauled by implementing a strict cross-well 
validation strategy. A new dataset split was generated at the well level: of the sixteen wells, fourteen were assigned to the training partition, and the remaining two wells\,---\,entirely absent from training\,---\,were reserved exclusively for testing. This ensured that the test set consisted of samples from geological wells unrepresented in the training data (Fig.~\ref{fig:cores_ds_3}). The class 
distributions for the resulting training and testing sets are illustrated in 
Fig.~\ref{fig:distrib_1} and Fig.~\ref{fig:distrib_2}, respectively.

\begin{figure}[!h]
    \centering
    \includegraphics[width=\linewidth,height=\textheight,keepaspectratio]{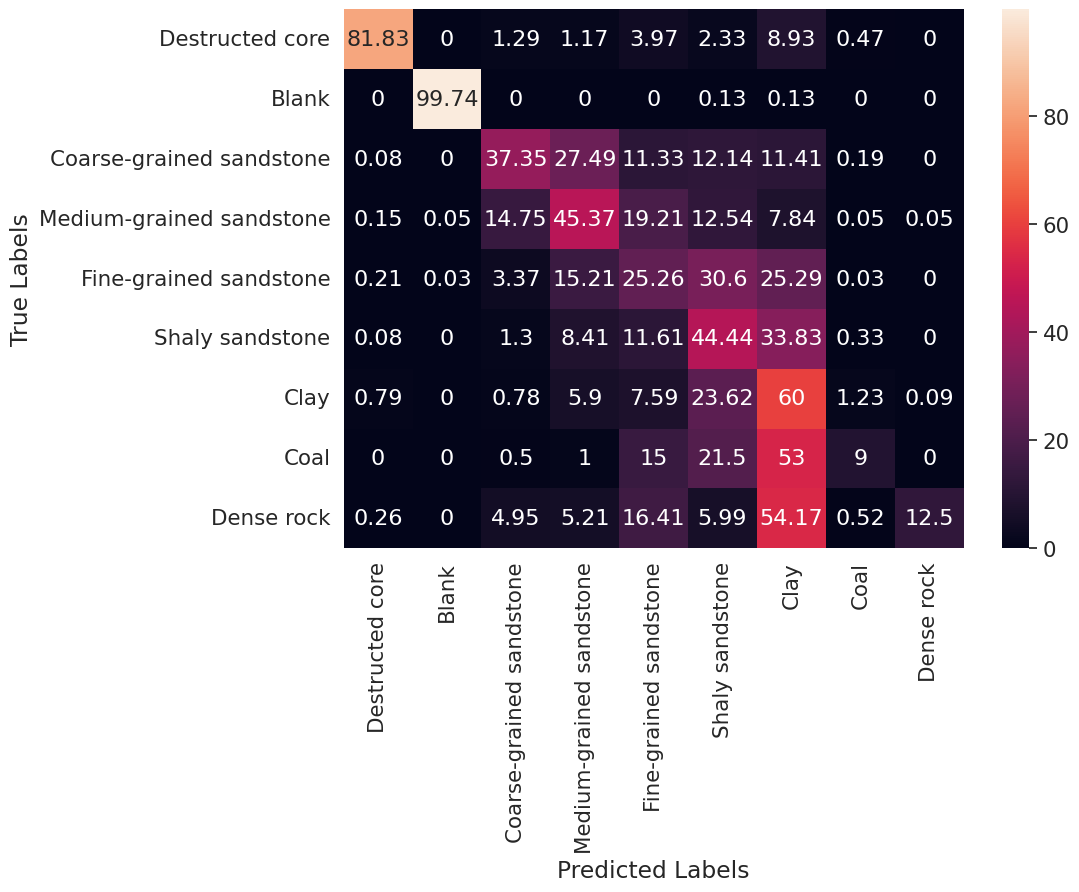}
    \caption{Confusion matrix of the DEP model using the strict cross-well test dataset. The results establish a realistic baseline, highlighting the difficulty of classifying minority facies (coal, dense rock).}
    \label{fig:cm_4}
\end{figure}

Under this robust validation framework, performance metrics converged to their true 
values. For the DEP backbone, the overall accuracy dropped to 50.8\%, with an MCA 
of 46.17\%. The corresponding confusion matrix (Fig.~\ref{fig:cm_4}) illustrated 
the inherent complexity of the task, particularly the systematic misclassification 
of rare minority classes — coal (0.2\% of training data) and dense rock (1.0\%) — 
as the dominant clay facies. This result of 50.8\% established an honest, unbiased 
baseline for automated lithofacies recognition in real-world geological 
applications.

\subsection{Architecture Benchmarking under Cross-Well Validation}
\label{sec:arch_benchmark}

\subsubsection{Vision Transformer (ViT)}

While the Vision Transformer (ViT) possesses a formidable capacity for global 
feature representation, it proved highly susceptible to severe overfitting within 
the data-constrained geological domain of this study. As detailed in 
Table~\ref{tab:vit_metrics}, the model rapidly achieved near-optimal training 
accuracy (93--98\%); however, its generalization capability degraded substantially 
during evaluation, yielding an overall test accuracy of only 48\% and a Mean Class 
Accuracy (MCA) of 37\%.

\begin{figure}[!h]
    \centering
    \includegraphics[width=\linewidth,height=\textheight,keepaspectratio]{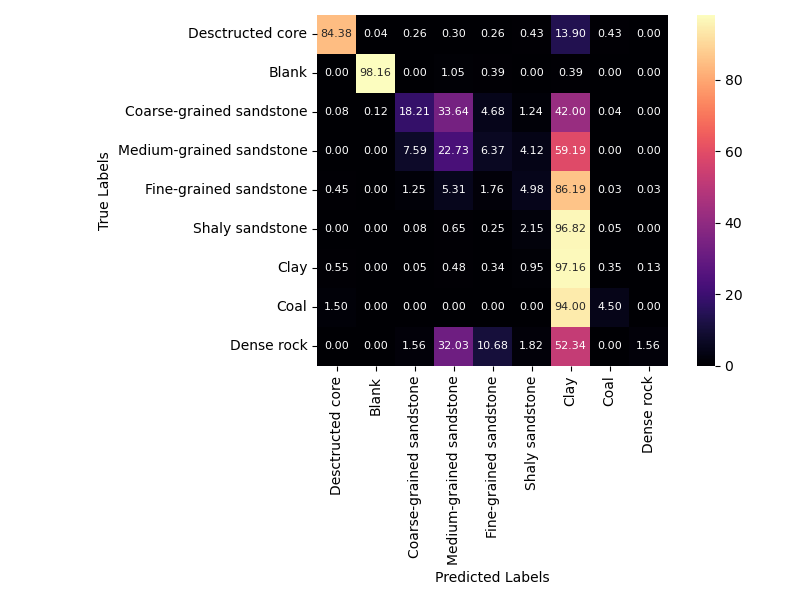}
    \caption{Confusion matrix of the Vision Transformer (ViT) on the 9-class cross-well test set, illustrating near-total collapse toward the dominant clay facies and poor generalization to minority classes.}
    \label{fig:cm_vit}
\end{figure}

This divergence constituted a textbook manifestation of the data-hungry nature of 
pure Transformer architectures. Lacking the built-in spatial inductive biases — 
such as translation invariance and local receptive fields — inherent to 
convolutional networks, the ViT effectively memorized the training set, including 
synthetic GAN artifacts and background noise, rather than learning generalizable 
lithological features. The confusion matrix (Fig.~\ref{fig:cm_vit} and 
Table~\ref{tab:cm_vit}) further illustrated this failure mode: the network exhibited 
severe model collapse toward the majority facies, predominantly classifying 
ambiguous samples as clay. Consequently, precision and recall for underrepresented 
classes — specifically coal and dense rock — were markedly low, confirming that the 
ViT architecture was unsuitable as a standalone classifier in data-scarce geological 
domains without domain-specific pre-training on large-scale lithological corpora.

\begin{table}[ht]
\centering
\scriptsize 
\setlength{\tabcolsep}{1.5pt} 

\caption{Vision Transformer confusion matrix for 9 classes.}
\label{tab:cm_vit}

\begin{tabularx}{\columnwidth}{@{} >{\raggedright\arraybackslash}X @{\hspace{1pt}} c c c c c c c c c @{}}
\toprule
\textbf{True $\backslash$ Pred} & 
\rotatebox{90}{\textbf{Destr.}} & 
\rotatebox{90}{\textbf{Blank}} & 
\rotatebox{90}{\textbf{C-gr.}} & 
\rotatebox{90}{\textbf{M-gr.}} & 
\rotatebox{90}{\textbf{F-gr.}} & 
\rotatebox{90}{\textbf{Shaly}} & 
\rotatebox{90}{\textbf{Clay}} & 
\rotatebox{90}{\textbf{Coal}} & 
\rotatebox{90}{\textbf{Dense}} \\
\midrule

Destructed core      & 1955 & 1   & 6   & 7   & 6   & 10  & 322  & 10 & 0 \\
Blank                & 0    & 746 & 0   & 8   & 3   & 0   & 3    & 0  & 0 \\
C-gr. sandstone      & 2    & 3   & 471 & 870 & 121 & 32  & 1086 & 1  & 0 \\
M-gr. sandstone      & 0    & 0   & 155 & 464 & 130 & 84  & 1208 & 0  & 0 \\
F-gr. sandstone      & 15   & 0   & 42  & 178 & 59  & 167 & 2890 & 1  & 1 \\
Shaly sandstone      & 0    & 0   & 3   & 26  & 10  & 86  & 3869 & 2  & 0 \\
Clay                 & 42   & 0   & 4   & 37  & 26  & 73  & 7479 & 27 & 10 \\
Coal                 & 3    & 0   & 0   & 0   & 0   & 0   & 188  & 9  & 0 \\
Dense rock           & 0    & 0   & 6   & 123 & 41  & 7   & 201  & 0  & 6 \\

\bottomrule
\end{tabularx}
\end{table}

\begin{table}[ht]
\centering
\caption{Per-class precision, recall, and F1-score for Vision Transformer (ViT) model on 9 core facies classes.}
\label{tab:vit_metrics}
\begin{tabular}{lcccc}
\toprule
\textbf{Class} & \textbf{Precision} & \textbf{Recall} & \textbf{F1-Score} & \textbf{Support} \\
\midrule
Destructed core           & 0.97 & 0.84 & 0.90 & 2317 \\
Blank                     & 0.99 & 0.98 & 0.99 & 760 \\
Coarse-grained sandstone  & 0.69 & 0.18 & 0.29 & 2586 \\
Medium-grained sandstone  & 0.27 & 0.23 & 0.25 & 2041 \\
Fine-grained sandstone    & 0.15 & 0.02 & 0.03 & 3353 \\
Shaly sandstone           & 0.19 & 0.02 & 0.04 & 3996 \\
Clay                      & 0.43 & 0.97 & 0.60 & 7698 \\
Coal                      & 0.18 & 0.04 & 0.07 & 200 \\
Dense rock                & 0.35 & 0.02 & 0.03 & 384 \\
\bottomrule
\end{tabular}
\end{table}

\subsubsection{Swin Tiny}

\begin{figure}[!h]
    \centering
    \includegraphics[width=\linewidth,height=\textheight,keepaspectratio]{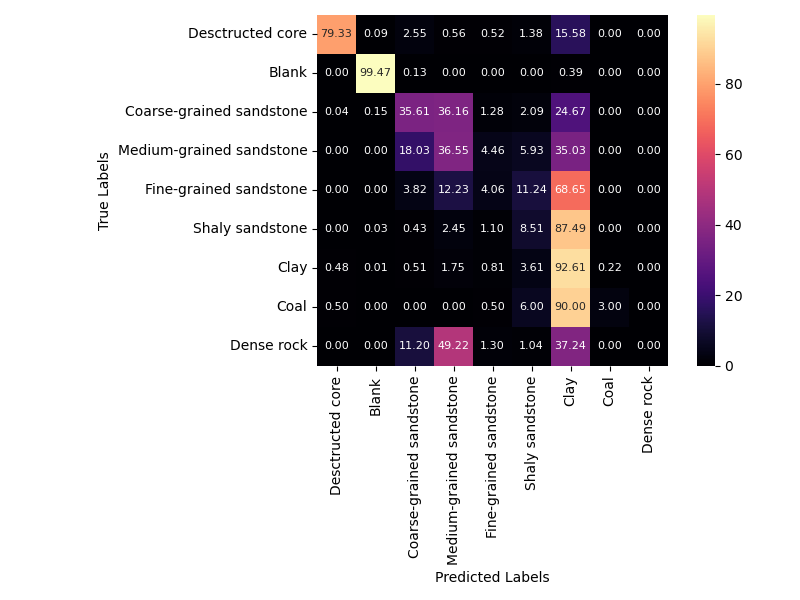}
    \caption{Confusion matrix of the Swin Tiny model on the 9-class 
    cross-well test set, illustrating severe bias toward the dominant 
    clay facies and near-zero recall for dense rock.}
    \label{fig:cm_swin_tiny}
\end{figure}

To address the lack of spatial inductive biases, the Swin Tiny architecture was 
evaluated as a complementary model to the pure Transformer approach. By employing 
shifted-window self-attention, Swin Tiny computes hierarchical feature 
representations that capture local spatial contexts at multiple scales. Despite 
these advantages, the baseline evaluation on the unbalanced dataset 
(Table~\ref{tab:swin_tiny_metrics}) revealed a persistent vulnerability to extreme 
class imbalance, with an overall accuracy of 51\% and an MCA of 39\%. As shown in 
the initial confusion matrix (Fig.~\ref{fig:cm_swin_tiny} and 
Table~\ref{tab:cm_swin_tiny}), the model exhibited a severe bias toward the dominant 
clay facies, with near-zero recall for minority classes such as dense rock.

To mitigate this limitation, a targeted data augmentation protocol was implemented 
exclusively for underrepresented classes. Minority samples were subjected to spatial 
and noise-based transformations — specifically, rotations of $\pm10^\circ$ and 
Gaussian noise with amplitude 10--30 — to expand intra-class variance. It should 
be noted that the performance figures reported in the consolidated comparison 
(Table~\ref{tab:model_comparison}) correspond to the \textit{unbalanced} 
configuration, whereas the results discussed in the remainder of this subsection 
reflect the \textit{balanced} configuration; these two settings are not directly 
comparable and address distinct operational objectives. Following the application 
of the balancing strategy, the model demonstrated a substantially improved capacity 
to recognize rare lithologies: recall for \textit{Coal} increased from 0.03 to 0.86 
and for \textit{Dense rock} from 0.00 to 0.71 
(Table~\ref{tab:swin_tiny_balanced_metrics}). The MCA consequently rose to 47\%, 
while the overall accuracy under the balanced configuration was 34\% — a reduction 
attributable to the deliberate reweighting of minority classes at the expense of 
majority-class discrimination.

\begin{figure}[!h]
    \centering
    \includegraphics[width=\linewidth,height=\textheight,keepaspectratio]{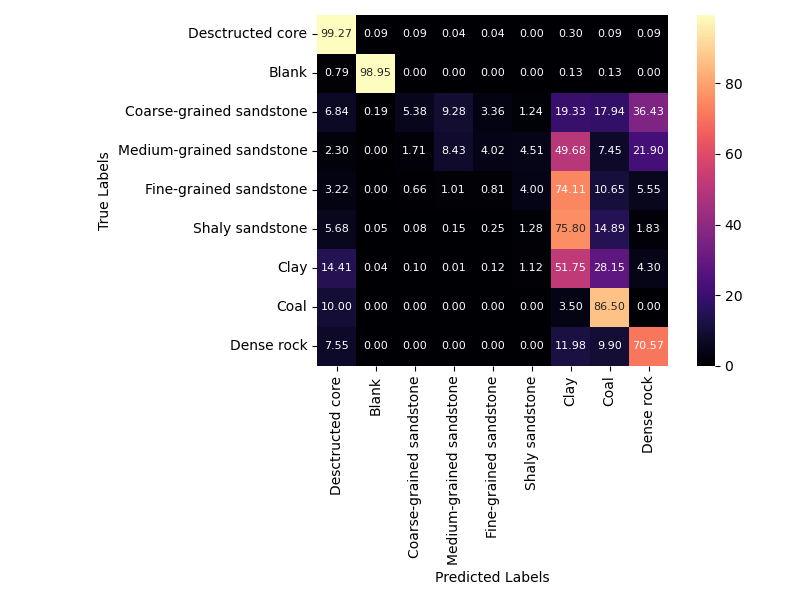}
    \caption{Confusion matrix of the Swin Tiny model after targeted 
    augmentation of minority classes, showing improved recall for coal 
    and dense rock at the cost of reduced accuracy for dominant facies.}
    \label{fig:cm_swin_tiny_balanced}
\end{figure}

However, this improvement came at a direct cost to majority-class performance. The 
aggressive rebalancing caused a marked decline in F1-score for dominant facies: 
\textit{Clay} dropped from 0.63 to 0.42, \textit{Coarse-grained sandstone} from 
0.44 to 0.10, and \textit{Medium-grained sandstone} from 0.33 to 0.14 
(Fig.~\ref{fig:cm_swin_tiny_balanced} and 
Table~\ref{tab:cm_swin_tiny_balanced}). This trade-off illustrated the fundamental 
tension between per-class sensitivity and global accuracy in highly imbalanced 
geological datasets, and motivated the confidence-based hybrid ensemble approach 
described in Section~\ref{sec:ensemble}.

\begin{table}[ht]
\centering
\caption{Per-class classification metrics for Swin Tiny model before dataset balancing.}
\label{tab:swin_tiny_metrics}
\begin{tabular}{lcccc}
\toprule
\textbf{Class} & \textbf{Precision} & \textbf{Recall} & \textbf{F1-Score} & \textbf{Support} \\
\midrule
Destructed core           & 0.98 & 0.79 & 0.88 & 2317 \\
Blank                     & 0.99 & 0.99 & 0.99 & 760 \\
Coarse-grained sandstone  & 0.58 & 0.36 & 0.44 & 2586 \\
Medium-grained sandstone  & 0.30 & 0.37 & 0.33 & 2041 \\
Fine-grained sandstone    & 0.35 & 0.04 & 0.07 & 3353 \\
Shaly sandstone           & 0.28 & 0.09 & 0.13 & 3996 \\
Clay                      & 0.48 & 0.93 & 0.63 & 7698 \\
Coal                      & 0.26 & 0.03 & 0.05 & 200 \\
Dense rock                & 0.00 & 0.00 & 0.00 & 384 \\
\bottomrule
\end{tabular}
\end{table}

\begin{table}[ht]
\centering
\caption{Per-class classification metrics for Swin Tiny model after dataset balancing.}
\label{tab:swin_tiny_balanced_metrics}
\begin{tabular}{lcccc}
\toprule
\textbf{Class} & \textbf{Precision} & \textbf{Recall} & \textbf{F1-Score} & \textbf{Support} \\
\midrule
Destructed core           & 0.57 & 0.99 & 0.73 & 2317 \\
Blank                     & 0.98 & 0.99 & 0.99 & 760 \\
Coarse-grained sandstone  & 0.67 & 0.05 & 0.10 & 2586 \\
Medium-grained sandstone  & 0.38 & 0.08 & 0.14 & 2041 \\
Fine-grained sandstone    & 0.12 & 0.01 & 0.02 & 3353 \\
Shaly sandstone           & 0.13 & 0.01 & 0.02 & 3996 \\
Clay                      & 0.36 & 0.52 & 0.42 & 7698 \\
Coal                      & 0.04 & 0.86 & 0.08 & 200 \\
Dense rock                & 0.12 & 0.71 & 0.21 & 384 \\
\bottomrule
\end{tabular}
\end{table}

\begin{table}[ht]
\centering
\scriptsize 
\setlength{\tabcolsep}{1.5pt}

\caption{Confusion matrix for Swin Tiny model for 9 classes.}
\label{tab:cm_swin_tiny}

\begin{tabularx}{\columnwidth}{@{} >{\raggedright\arraybackslash}X c c c c c c c c c @{}}
\toprule
\textbf{True $\backslash$ Pred} & 
\rotatebox{90}{\textbf{Destr.}} & 
\rotatebox{90}{\textbf{Blank}} & 
\rotatebox{90}{\textbf{C-gr.}} & 
\rotatebox{90}{\textbf{M-gr.}} & 
\rotatebox{90}{\textbf{F-gr.}} & 
\rotatebox{90}{\textbf{Shaly}} & 
\rotatebox{90}{\textbf{Clay}} & 
\rotatebox{90}{\textbf{Coal}} & 
\rotatebox{90}{\textbf{Dense}} \\
\midrule

Destructed core      & 1838 & 2   & 59  & 13  & 12  & 32  & 361  & 0  & 0 \\
Blank                & 0    & 756 & 1   & 0   & 0   & 0   & 3    & 0  & 0 \\
C-gr. sandstone      & 1    & 4   & 921 & 935 & 33  & 54  & 638  & 0  & 0 \\
M-gr. sandstone      & 0    & 0   & 368 & 746 & 91  & 121 & 715  & 0  & 0 \\
F-gr. sandstone      & 0    & 0   & 128 & 410 & 136 & 377 & 2302 & 0  & 0 \\
Shaly sandstone      & 0    & 1   & 17  & 98  & 44  & 340 & 3496 & 0  & 0 \\
Clay                 & 37   & 1   & 39  & 135 & 62  & 278 & 7129 & 17 & 0 \\
Coal                 & 1    & 0   & 0   & 0   & 1   & 12  & 180  & 6  & 0 \\
Dense rock           & 0    & 0   & 43  & 189 & 5   & 4   & 143  & 0  & 0 \\

\bottomrule
\end{tabularx}
\end{table}

\begin{table}[ht]
\centering
\scriptsize 
\setlength{\tabcolsep}{1.5pt} 

\caption{Confusion matrix for Swin Tiny (balanced dataset).}
\label{tab:cm_swin_tiny_balanced}

\begin{tabularx}{\columnwidth}{@{} >{\raggedright\arraybackslash}X c c c c c c c c c @{}}
\toprule
\textbf{True $\backslash$ Pred} & 
\rotatebox{90}{\textbf{Destr.}} & 
\rotatebox{90}{\textbf{Blank}} & 
\rotatebox{90}{\textbf{C-gr.}} & 
\rotatebox{90}{\textbf{M-gr.}} & 
\rotatebox{90}{\textbf{F-gr.}} & 
\rotatebox{90}{\textbf{Shaly}} & 
\rotatebox{90}{\textbf{Clay}} & 
\rotatebox{90}{\textbf{Coal}} & 
\rotatebox{90}{\textbf{Dense}} \\
\midrule

Destructed core  & 2300 & 2   & 2   & 1   & 1   & 0   & 7    & 2    & 2 \\
Blank            & 6    & 752 & 0   & 0   & 0   & 0   & 1    & 1    & 0 \\
C-gr. sandstone  & 177  & 5   & 139 & 240 & 87  & 32  & 500  & 464  & 942 \\
M-gr. sandstone  & 47   & 0   & 35  & 172 & 82  & 92  & 1014 & 152  & 447 \\
F-gr. sandstone  & 108  & 0   & 22  & 34  & 27  & 134 & 2485 & 357  & 186 \\
Shaly sandstone  & 227  & 2   & 3   & 6   & 10  & 51  & 3029 & 595  & 73 \\
Clay             & 1109 & 3   & 8   & 1   & 9   & 86  & 3984 & 2167 & 331 \\
Coal             & 20   & 0   & 0   & 0   & 0   & 0   & 7    & 173  & 0 \\
Dense rock       & 29   & 0   & 0   & 0   & 0   & 0   & 46   & 38   & 271 \\

\bottomrule
\end{tabularx}
\end{table}

\subsubsection{EfficientNetV2}

Among the evaluated standalone architectures, EfficientNetV2 demonstrated the most 
robust baseline generalization, achieving an overall accuracy of 53\% and an MCA 
of 48\% (Table~\ref{tab:efficient_net_v2_metrics}). This relative stability was 
attributed to the architecture's compound scaling methodology, which systematically 
optimizes network depth, width, and input resolution simultaneously. This design 
allowed the network to efficiently extract multi-scale textural features while 
resisting the severe overfitting observed in Transformer-based models under the 
data-constrained conditions of this study.

Despite these strengths, the confusion matrix (Fig.~\ref{fig:ablation_cm_comp}b 
and Table~\ref{tab:cm_efficient_net_v2}) revealed that the model remained 
constrained by the inherent visual ambiguity of core samples, particularly the 
systematic confusion between fine-grained and shaly sandstone variants — facies 
that share overlapping grain-scale textural signatures and are therefore 
intrinsically difficult to discriminate using RGB imagery alone.

\begin{table}[ht]
\centering
\caption{Per-class classification metrics for EfficientNetV2 model on 9 lithological facies.}
\label{tab:efficient_net_v2_metrics}
\begin{tabular}{lcccc}
\toprule
\textbf{Class} & \textbf{Precision} & \textbf{Recall} & \textbf{F1-Score} & \textbf{Support} \\
\midrule
Destructed core           & 0.95 & 0.78 & 0.85 & 2317 \\
Blank                     & 0.99 & 1.00 & 0.99 & 760 \\
Coarse-grained sandstone  & 0.75 & 0.47 & 0.58 & 2586 \\
Medium-grained sandstone  & 0.36 & 0.61 & 0.45 & 2041 \\
Fine-grained sandstone    & 0.34 & 0.16 & 0.22 & 3353 \\
Shaly sandstone           & 0.31 & 0.39 & 0.34 & 3996 \\
Clay                      & 0.58 & 0.66 & 0.62 & 7698 \\
Coal                      & 0.27 & 0.18 & 0.21 & 200 \\
Dense rock                & 0.37 & 0.07 & 0.11 & 384 \\
\bottomrule
\end{tabular}
\end{table}

\begin{table}[ht]
\centering
\scriptsize 
\setlength{\tabcolsep}{1.5pt} 

\caption{Confusion matrix for EfficientNetV2 model.}
\label{tab:cm_efficient_net_v2}

\begin{tabularx}{\columnwidth}{@{} >{\raggedright\arraybackslash}X c c c c c c c c c @{}}
\toprule
\textbf{True $\backslash$ Pred} & 
\rotatebox{90}{\textbf{Destr.}} & 
\rotatebox{90}{\textbf{Blank}} & 
\rotatebox{90}{\textbf{C-gr.}} & 
\rotatebox{90}{\textbf{M-gr.}} & 
\rotatebox{90}{\textbf{F-gr.}} & 
\rotatebox{90}{\textbf{Shaly}} & 
\rotatebox{90}{\textbf{Clay}} & 
\rotatebox{90}{\textbf{Coal}} & 
\rotatebox{90}{\textbf{Dense}} \\
\midrule

Destructed core      & 1798 & 2   & 91   & 18   & 52  & 4    & 346  & 6  & 0 \\
Blank                & 0    & 758 & 1    & 0    & 0   & 1    & 0    & 0  & 0 \\
C-gr. sandstone      & 11   & 3   & 1217 & 748  & 118 & 222  & 263  & 4  & 0 \\
M-gr. sandstone      & 3    & 1   & 205  & 1235 & 237 & 205  & 155  & 0  & 0 \\
F-gr. sandstone      & 32   & 0   & 76   & 654  & 528 & 1262 & 796  & 0  & 5 \\
Shaly sandstone      & 6    & 0   & 11   & 320  & 226 & 1572 & 1853 & 6  & 2 \\
Clay                 & 43   & 4   & 7    & 311  & 330 & 1834 & 5051 & 82 & 36 \\
Coal                 & 5    & 0   & 0    & 0    & 0   & 8    & 151  & 36 & 0 \\
Dense rock           & 0    & 0   & 8    & 186  & 57  & 15   & 92   & 1  & 25 \\

\bottomrule
\end{tabularx}
\end{table}

\subsubsection{ConvNeXt}
\begin{figure}[!h]
    \centering
    \includegraphics[width=\linewidth,height=\textheight,keepaspectratio]{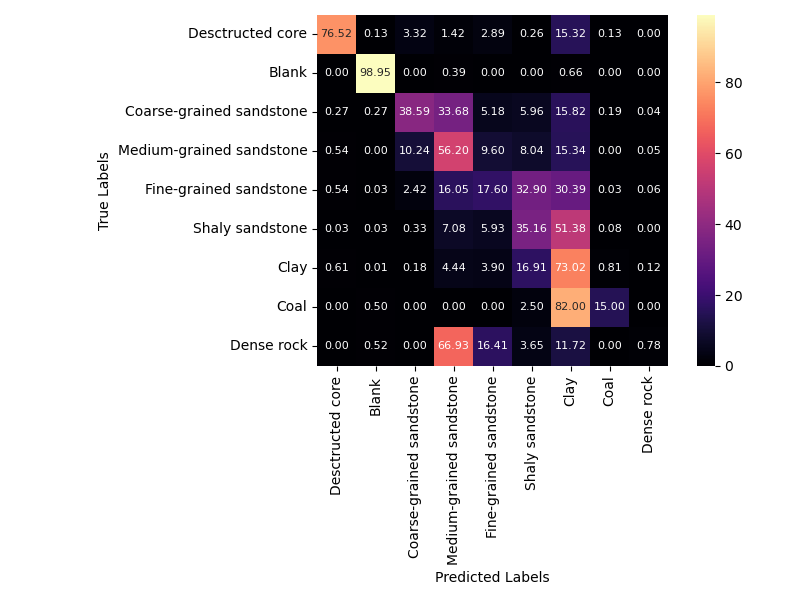}
    \caption{Confusion matrix of the ConvNeXt model on 9 lithological facies under 
    cross-well validation (overall accuracy: 53\%). While global performance is 
    comparable to EfficientNetV2, ConvNeXt exhibits a more severe sensitivity deficit 
    toward extreme minority classes: dense rock achieved a recall of only 1\%, 
    indicating that Transformer-inspired architectural modifications do not 
    compensate for the absence of compound scaling under severe class imbalance.}
    \label{fig:cm_conv_next}
\end{figure}

ConvNeXt achieved an overall accuracy of 53\% and an MCA of 46\% 
(Table~\ref{tab:conv_next}), closely matching EfficientNetV2 and confirming that 
modernized CNN architectures provide a stable baseline under the data-constrained 
conditions of this study. However, the confusion matrix 
(Fig.~\ref{fig:cm_conv_next} and Table~\ref{tab:cm_conv_next}) revealed a severe 
sensitivity deficit for extreme minority classes — most notably dense rock, with a 
recall of only 1\%. The convergence of both CNN architectures at the same accuracy 
ceiling suggests a dataset-level limitation shared across model families.

\begin{table}[ht]
\centering
\caption{Per-class classification metrics for ConvNeXt model on 9 lithological facies.}
\label{tab:conv_next}
\begin{tabular}{lcccc}
\toprule
\textbf{Class} & \textbf{Precision} & \textbf{Recall} & \textbf{F1-Score} & \textbf{Support} \\
\midrule
Destructed core           & 0.95 & 0.77 & 0.85 & 2317 \\
Blank                     & 0.98 & 0.99 & 0.98 & 760 \\
Coarse-grained sandstone  & 0.72 & 0.39 & 0.50 & 2586 \\
Medium-grained sandstone  & 0.33 & 0.56 & 0.42 & 2041 \\
Fine-grained sandstone    & 0.37 & 0.18 & 0.24 & 3353 \\
Shaly sandstone           & 0.34 & 0.35 & 0.34 & 3996 \\
Clay                      & 0.56 & 0.73 & 0.64 & 7698 \\
Coal                      & 0.29 & 0.15 & 0.20 & 200 \\
Dense rock                & 0.19 & 0.01 & 0.01 & 384 \\
\bottomrule
\end{tabular}
\end{table}

\begin{table}[ht]
\centering
\scriptsize 
\setlength{\tabcolsep}{1.5pt} 

\caption{Confusion matrix for ConvNeXt model.}
\label{tab:cm_conv_next}

\begin{tabularx}{\columnwidth}{@{} >{\raggedright\arraybackslash}X c c c c c c c c c @{}}
\toprule
\textbf{True $\backslash$ Pred} & 
\rotatebox{90}{\textbf{Destr.}} & 
\rotatebox{90}{\textbf{Blank}} & 
\rotatebox{90}{\textbf{C-gr.}} & 
\rotatebox{90}{\textbf{M-gr.}} & 
\rotatebox{90}{\textbf{F-gr.}} & 
\rotatebox{90}{\textbf{Shaly}} & 
\rotatebox{90}{\textbf{Clay}} & 
\rotatebox{90}{\textbf{Coal}} & 
\rotatebox{90}{\textbf{Dense}} \\
\midrule

Destructed core      & 1773 & 3   & 77  & 33   & 67  & 6    & 355  & 3  & 0 \\
Blank                & 0    & 752 & 0   & 3    & 0   & 0    & 5    & 0  & 0 \\
C-gr. sandstone      & 7    & 7   & 998 & 871  & 134 & 154  & 409  & 5  & 1 \\
M-gr. sandstone      & 11   & 0   & 209 & 1147 & 196 & 164  & 313  & 0  & 1 \\
F-gr. sandstone      & 18   & 1   & 81  & 538  & 590 & 1103 & 1019 & 1  & 2 \\
Shaly sandstone      & 1    & 1   & 13  & 283  & 237 & 1405 & 2053 & 3  & 0 \\
Clay                 & 47   & 1   & 14  & 342  & 300 & 1302 & 5621 & 62 & 9 \\
Coal                 & 0    & 1   & 0   & 0    & 0   & 5    & 164  & 30 & 0 \\
Dense rock           & 0    & 2   & 0   & 257  & 63  & 14   & 45   & 0  & 3 \\

\bottomrule
\end{tabularx}
\end{table}

\subsubsection{Ablation Study: Impact of GAN Reconstruction}
\label{sec:ablation_analysis}

To evaluate the necessity of the GAN-based reconstruction stage, we compared two 
EfficientNetV2 configurations under identical cross-well validation: (i)~a baseline 
operating on raw imagery with physical defects, and (ii)~the full pipeline with 
CRA-GAN reconstruction. Per-class results are summarized in 
Table~\ref{tab:ablation_comparison}, Table~\ref{tab:baseline_detailed_metrics}, and Table~\ref{tab:cm_baseline_raw}.

Both configurations achieved 53\% overall accuracy, indicating that reconstruction 
did not produce a uniform classification gain despite substantially improving 
perceptual quality metrics.

\begin{figure*}[t!]
    \centering
    \begin{subfigure}{0.48\textwidth}
        \centering
        \includegraphics[width=\linewidth]{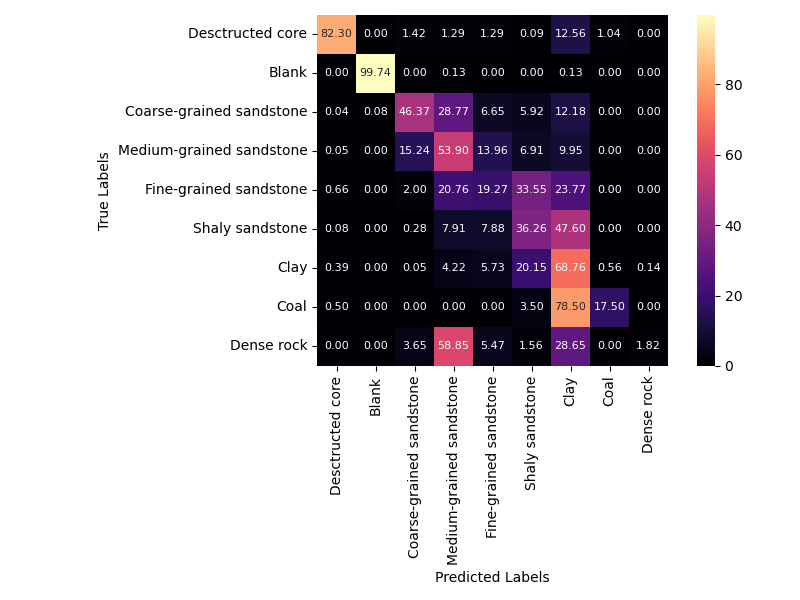} 
        \caption{Performance on raw data (original damaged imagery).}
    \end{subfigure}
    \hfill
    \begin{subfigure}{0.48\textwidth}
        \centering
        \includegraphics[width=\linewidth]{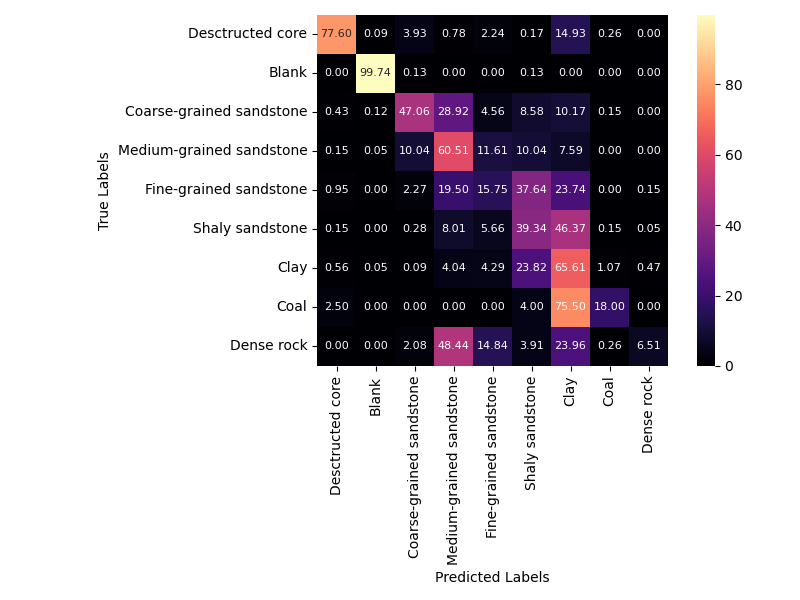}
        \caption{Performance on reconstructed data (after GAN-based inpainting).}
    \end{subfigure}
    \caption{Visual comparison of confusion matrices for the EfficientNet-V2 model.}
    \label{fig:ablation_cm_comp}
\end{figure*}

Per-class analysis, however, revealed targeted improvements (Fig.~\ref{fig:ablation_cm_comp}). For the geologically 
critical \textit{Dense rock} class, recall improved from 0.02 to 0.07 — a 
3.5-fold increase — while \textit{Medium-grained sandstone} F1-score improved 
by $+$0.05. These gains indicate that CRA-GAN restored diagnostic grain-scale 
textures in lithologies where structural continuity was essential for feature 
extraction. Conversely, for classes with strong textural regularity (\textit{Clay}, 
\textit{Blank}), reconstruction introduced marginal noise with negligible or 
slightly negative effects on F1-score.

CRA-GAN therefore functions as a targeted preprocessing stage: its benefit is 
concentrated in structurally complex, underrepresented facies rather than acting 
as a universal accuracy enhancer.

\begin{table}[ht]
\centering
\caption{Quantitative Impact of GAN Reconstruction on Classification Performance.}
\label{tab:ablation_comparison}
\begin{tabular}{lcccc}
\toprule
\textbf{Class} & \textbf{Raw data (F1)} & \textbf{GAN data(F1)} & \textbf{$\Delta$} \\
\midrule
Destructed core           & 0.89 & 0.85 & -0.04 \\
Blank                     & 1.00 & 0.99 & -0.01 \\
Coarse-grained sandstone  & 0.57 & 0.58 & +0.01 \\
Medium-grained sandstone  & 0.40 & 0.45 & +0.05 \\
Fine-grained sandstone    & 0.25 & 0.22 & -0.03 \\
Shaly sandstone           & 0.34 & 0.34 & 0.00 \\
Clay                      & 0.63 & 0.62 & -0.01 \\
Coal                      & 0.23 & 0.21 & -0.02 \\
Dense rock                & 0.03 & 0.11 & +0.08 \\
\bottomrule
\end{tabular}
\end{table}

\begin{table}[ht]
\centering
\caption{Per-class classification metrics for EfficientNetV2
model on 9 lithological facies (no Inpainting).}
\label{tab:baseline_detailed_metrics}
\begin{tabular}{lcccc}
\toprule
\textbf{Class} & \textbf{Precision} & \textbf{Recall} & \textbf{F1-Score} & \textbf{Support} \\
\midrule
Destructed core           & 0.97 & 0.82 & 0.89 & 2317 \\
Blank                     & 1.00 & 1.00 & 1.00 & 760 \\
Coarse-grained sandstone  & 0.73 & 0.46 & 0.57 & 2586 \\
Medium-grained sandstone  & 0.32 & 0.54 & 0.40 & 2041 \\
Fine-grained sandstone    & 0.34 & 0.19 & 0.25 & 3353 \\
Shaly sandstone           & 0.33 & 0.36 & 0.34 & 3996 \\
Clay                      & 0.58 & 0.69 & 0.63 & 7698 \\
Coal                      & 0.34 & 0.17 & 0.23 & 200 \\
Dense rock                & 0.39 & 0.02 & 0.03 & 384 \\
\bottomrule

\end{tabular}
\end{table}

\begin{table}[ht]
\centering
\scriptsize 
\setlength{\tabcolsep}{1.5pt} 

\caption{Numerical confusion matrix for EfficientNetV2 (no Inpainting).}
\label{tab:cm_baseline_raw}

\begin{tabularx}{\columnwidth}{@{} >{\raggedright\arraybackslash}X c c c c c c c c c @{}}
\toprule
\textbf{True $\backslash$ Pred} & 
\rotatebox{90}{\textbf{Destr.}} & 
\rotatebox{90}{\textbf{Blank}} & 
\rotatebox{90}{\textbf{C-gr.}} & 
\rotatebox{90}{\textbf{M-gr.}} & 
\rotatebox{90}{\textbf{F-gr.}} & 
\rotatebox{90}{\textbf{Shaly}} & 
\rotatebox{90}{\textbf{Clay}} & 
\rotatebox{90}{\textbf{Coal}} & 
\rotatebox{90}{\textbf{Dense}} \\
\midrule

Destructed core      & 1907 & 0   & 33  & 30   & 30   & 2    & 291  & 24  & 0   \\
Blank                & 0    & 758 & 0   & 1    & 0    & 0    & 1    & 0   & 0   \\
C-gr. sandstone      & 1    & 2   & 1199 & 744  & 172  & 153  & 315  & 0   & 0   \\
M-gr. sandstone      & 1    & 0   & 311 & 1100 & 285  & 141  & 203  & 0   & 0   \\
F-gr. sandstone      & 22   & 0   & 67  & 696  & 646  & 1125 & 797  & 0   & 0   \\
Shaly sandstone      & 3    & 0   & 11  & 316  & 315  & 1449 & 1902 & 0   & 0   \\
Clay                 & 30   & 0   & 4   & 325  & 441  & 1551 & 5293 & 43  & 11  \\
Coal                 & 1    & 0   & 0   & 0    & 0    & 7    & 157  & 35  & 0   \\
Dense rock           & 0    & 0   & 14  & 226  & 21   & 6    & 110  & 0   & 7   \\

\bottomrule
\end{tabularx}
\end{table}

\subsubsection{Hybrid Ensemble: EfficientNetV2 + Swin Tiny}
\label{sec:ensemble}

To capitalize on architectural synergies, a confidence-based hybrid ensemble 
framework was developed. EfficientNetV2 served as the primary predictor, but was 
conditionally overridden by the Swin Tiny model for four target classes (blank, 
destructed core, coal, and dense rock) when the Transformer's softmax confidence 
exceeded an 80\% threshold.

The four override classes\,---\,blank, destructed core, coal, and dense
rock\,---\,were selected by comparing the standalone validation-set confusion
matrices of the two models. On these categories, Swin Tiny substantially
outperformed EfficientNetV2 in per-class accuracy\,---\,destructed core 94.76\% vs.\
74.26\%, blank 93.50\% vs.\ 91.06\%, coal 61.54\% vs.\ 13.46\%, and dense rock
97.92\% vs.\ 10.42\%\,---\,whereas EfficientNetV2 remained the stronger predictor
for the high-frequency sandstone and clay facies that dominate the dataset. The
Transformer's global self-attention captured the large-scale, low-frequency
structural cues that characterize these classes, which explains its marked
advantage on the economically critical coal and dense-rock facies. Restricting the
override to these four classes therefore exploits each model where it is
demonstrably stronger, rather than overriding indiscriminately. The 80\%
softmax-confidence threshold was adopted as a deliberately conservative operating
point, intended to trigger a Swin Tiny override only when the Transformer is highly
confident in its prediction. This high-confidence requirement limits the risk that
an uncertain minority-class prediction displaces a correct majority-class one. Both
the override classes and the confidence threshold were determined exclusively on
the validation set, without reference to the held-out test wells; the test-set MCA
improvement reported below therefore reflects genuine generalization rather than
test-set optimization. The threshold was not subjected to an exhaustive search, and
a systematic sensitivity analysis of this operating point is identified as a
direction for future refinement.

\begin{figure}[!h]
    \centering
    \includegraphics[width=\linewidth,height=\textheight,keepaspectratio]{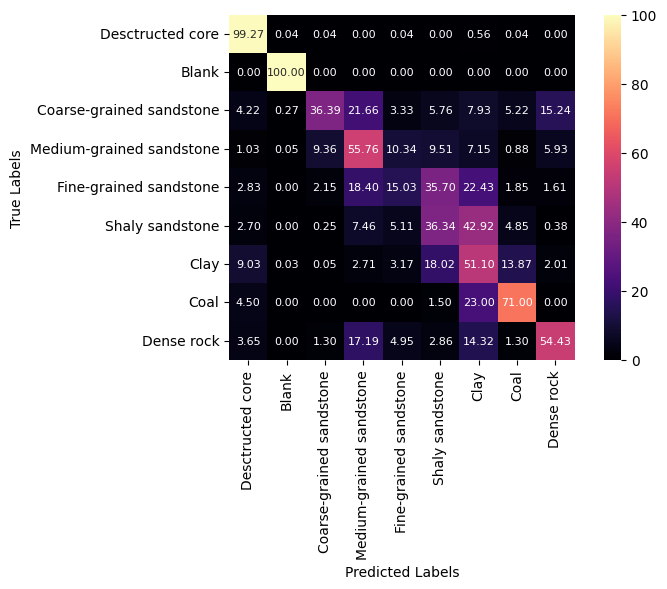}
    \caption{Confusion matrix of the hybrid EfficientNetV2 + Swin Tiny 
    ensemble on the 9-class cross-well test set, demonstrating improved 
    Mean Class Accuracy (MCA=58\%) through confidence-based selective override 
    of minority class predictions.}
    \label{fig:cm_eff_net_v2_and_swin_tiny}
\end{figure}

As illustrated in the ensemble confusion matrix 
(Fig.~\ref{fig:cm_eff_net_v2_and_swin_tiny} and 
Table~\ref{tab:efficientnetv2_swin_tiny}), the hybrid framework effectively 
compensated for the individual failure modes of each constituent model. By 
selectively overriding predictions for structurally ambiguous minority classes, 
the ensemble achieved an overall accuracy of 49\% while raising the Mean Class 
Accuracy (MCA) to 58\% — a gain of 10~percentage points over the standalone 
EfficientNetV2 baseline.

The reduction in overall accuracy from 53\% to 49\% warrants explicit 
justification. In a balanced dataset, overall accuracy and MCA converge and are 
equally informative; however, under the severe class imbalance present in this 
study — where clay alone accounted for approximately 32\% of the test set — 
overall accuracy was disproportionately governed by majority-class performance 
and therefore constituted a misleading indicator of practical utility. MCA, by 
contrast, assigns equal weight to each lithofacies irrespective of its frequency, 
providing a more faithful measure of a model's ability to characterize the full 
lithological column. The observed 4-percentage-point decline in overall accuracy 
reflected a deliberate rebalancing of the sensitivity--specificity trade-off: the 
ensemble sacrificed a marginal fraction of majority-class correct predictions in 
order to substantially improve the detection of geologically and economically 
significant minority lithofacies. Specifically, recall for \textit{Coal} and 
\textit{Dense rock} — lithofacies whose misclassification carries direct 
consequences for reservoir characterization and resource estimation — improved 
markedly under the ensemble configuration. In practical geological workflows, 
the cost of failing to identify a coal seam or a dense impermeable layer is 
considerably greater than a marginal reduction in aggregate classification rate; 
the ensemble is therefore the recommended configuration for laboratory-grade 
lithological analysis, as discussed in Section~\ref{sec:deployment}.

This result further demonstrated that the persistent gap between global accuracy 
and MCA could be partially bridged by combining the local texture sensitivity of 
CNNs with the long-range contextual reasoning of Transformers, without sacrificing 
computational tractability.
\begin{table}[ht]
\centering
\scriptsize 
\setlength{\tabcolsep}{1.5pt} 

\caption{Confusion matrix for EfficientNetV2 + Swin Tiny (confidence-based ensemble).}
\label{tab:efficientnetv2_swin_tiny}

\begin{tabularx}{\columnwidth}{@{} >{\raggedright\arraybackslash}X c c c c c c c c c @{}}
\toprule
\textbf{True $\backslash$ Pred} & 
\rotatebox{90}{\textbf{Destr.}} & 
\rotatebox{90}{\textbf{Blank}} & 
\rotatebox{90}{\textbf{C-gr.}} & 
\rotatebox{90}{\textbf{M-gr.}} & 
\rotatebox{90}{\textbf{F-gr.}} & 
\rotatebox{90}{\textbf{Shaly}} & 
\rotatebox{90}{\textbf{Clay}} & 
\rotatebox{90}{\textbf{Coal}} & 
\rotatebox{90}{\textbf{Dense}} \\
\midrule

Destructed core      & 2300 & 1   & 1   & 0     & 1   & 0     & 13   & 1    & 0 \\
Blank                & 0    & 760 & 0   & 0     & 0   & 0     & 0    & 0    & 0 \\
C-gr. sandstone      & 109  & 7   & 941 & 560   & 86  & 149   & 205  & 135  & 394 \\
M-gr. sandstone      & 21   & 1   & 191 & 1138  & 211 & 194   & 146  & 18   & 121 \\
F-gr. sandstone      & 95   & 0   & 72  & 617   & 504 & 1197  & 752  & 62   & 54 \\
Shaly sandstone      & 108  & 0   & 10  & 298   & 204 & 1452  & 1715 & 194  & 15 \\
Clay                 & 695  & 2   & 4   & 209   & 244 & 1387  & 3934 & 1068 & 155 \\
Coal                 & 9    & 0   & 0   & 0     & 0   & 3     & 46   & 142  & 0 \\
Dense rock           & 14   & 0   & 5   & 66    & 19  & 11    & 55   & 5   & 209 \\

\bottomrule
\end{tabularx}
\end{table}

\begin{figure}[ht]
    \centering
    \includegraphics[width=\columnwidth]{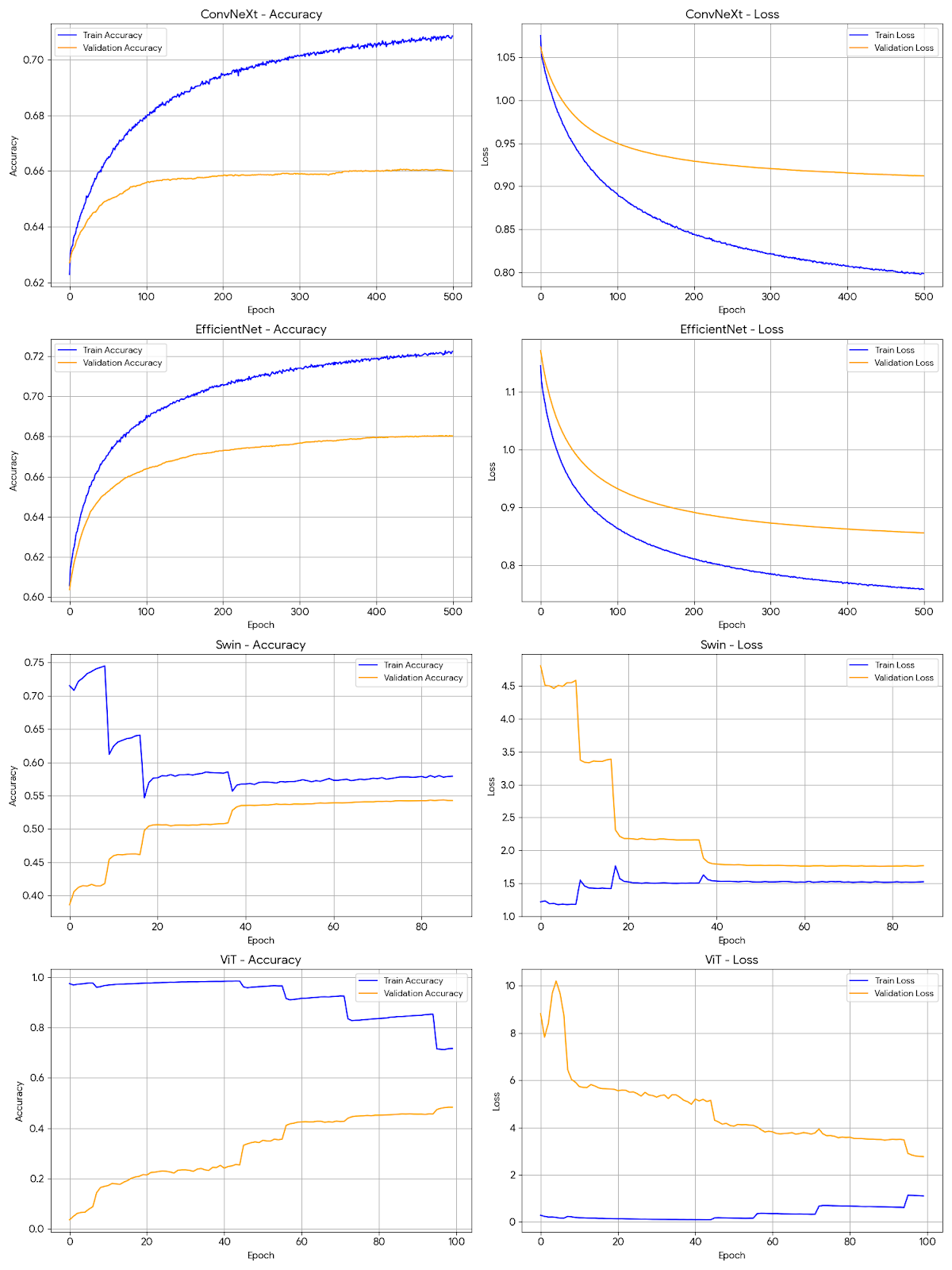} 
    \caption{Training and validation convergence curves (accuracy and loss) for the evaluated classification models.}
    \label{fig:convergence_curves_classifier}
\end{figure}

\begin{table}[ht]
\centering
\scriptsize
\caption{Performance comparison of core classification models for 9 classes with selected test wells. Results are reported with 95\% confidence intervals (calculated via 1,000 bootstrap iterations for deep models and analytically for the baseline).}
\label{tab:model_comparison}
\begin{tabularx}{\columnwidth}{@{} >{\raggedright\arraybackslash}X c c @{}}
\toprule
\textbf{Model} & \textbf{Accuracy (\%)} & \textbf{Mean Class Accuracy (\%)} \\ 
\midrule
DEP                            & 50.80 $\pm$ 0.64 & 46.17 $\pm$ 0.76 \\
Vision Transformer (ViT)       & 48.00 $\pm$ 0.64 & 36.00 $\pm$ 0.68 \\
Swin Tiny                      & 51.00 $\pm$ 0.66 & 39.00 $\pm$ 0.62 \\
EfficientNetV2 (No Inpainting) & 53.00 $\pm$ 0.66 & 48.00 $\pm$ 0.88 \\
EfficientNetV2                 & \textbf{53.00 $\pm$ 0.63} & 48.00 $\pm$ 0.90 \\
Ensemble (EffNetV2 + Swin Tiny)& 49.00 $\pm$ 0.68 & \textbf{58.00 $\pm$ 0.91} \\
ConvNeXt                       & 51.00 $\pm$ 0.64 & 45.00 $\pm$ 0.84 \\
\bottomrule
\end{tabularx}
\begin{tablenotes}
\footnotesize
\item Ensemble results correspond to the laboratory deployment 
scenario (Scenario~2, Section~\ref{sec:deployment}); 
overall accuracy reflects the deliberate sensitivity--specificity 
trade-off described in Section~\ref{sec:ensemble}.
\end{tablenotes}
\end{table}

\section{Discussion}
\label{sec:discussion}

The relationship between generative data augmentation and classification accuracy proved complex in this study. As noted by Likó and Kovács~\cite{Liko2024Augmentation}, while deep learning-based augmentation can improve general metrics, it requires meticulous handling to avoid introducing noise that degrades fine-grained classification. Our results confirmed this trade-off within the context of geological texture inpainting.

\subsection{Training Stability and Convergence}
\label{sec:stability}
The convergence profiles (Fig.~\ref{fig:convergence_curves_classifier}) revealed 
distinct learning dynamics across the evaluated architectures.

For CNN-based models (EfficientNetV2 and ConvNeXt), training and validation 
metrics plateaued concurrently without catastrophic divergence, indicating stable 
extraction of the available textural representations. Both architectures reached 
a performance ceiling of 51--53\%, consistent with the high degree of inter-class 
overlap visible in the t-SNE projection (Fig.~\ref{fig:tsne_comparison}).

The Vision Transformer (ViT) exhibited a contrasting pattern: training accuracy 
reached 93--98\%, while validation performance stagnated at substantially lower 
values. Without the spatial inductive biases of convolutional networks, ViT 
memorized the training distribution — including synthetic GAN artifacts — rather 
than learning transferable lithological features. This overfitting behavior 
confirmed that the dataset size was insufficient for pure self-attention 
architectures in this domain.

\subsection{The Semantic Gap in Texture Reconstruction}
\label{sec:tsne}

Despite high reconstruction fidelity (PSNR 28.7~dB, FID 74.01), classification 
accuracy plateaued at 53\%, revealing a divergence between perceptual quality and 
downstream utility. The CRA-GAN synthesized grain-scale details that satisfied 
distributional metrics but did not reliably preserve the lithological properties 
of the missing region — producing what we term "hallucinated" textures: 
statistically plausible features that lack class-discriminative value.

\begin{figure}[htbp]
    \centering
    \begin{minipage}{0.48\textwidth}
        \centering
        \includegraphics[width=\textwidth]{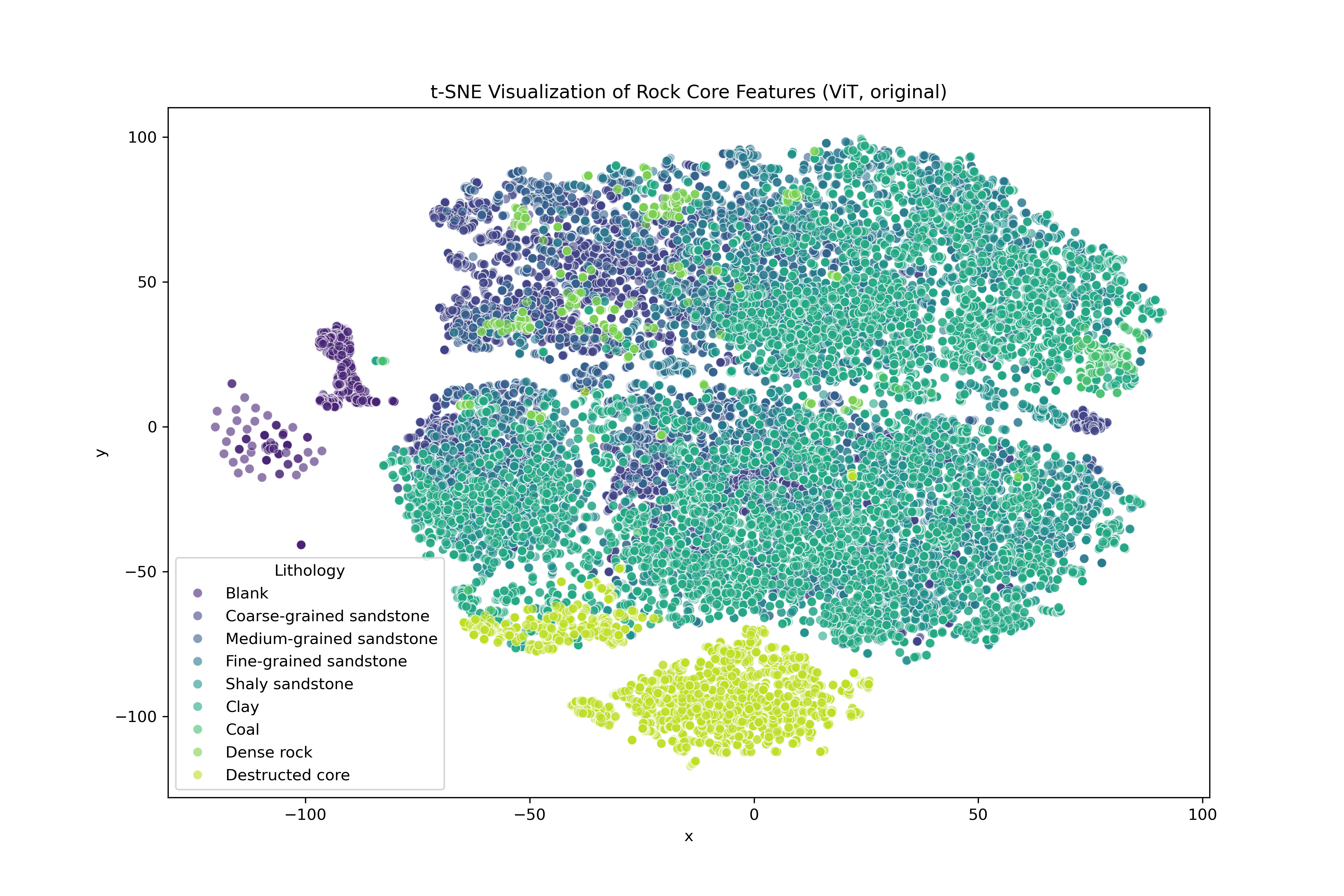}
        \caption*{(a) Original Images.}
    \end{minipage}
    \hfill
    \begin{minipage}{0.48\textwidth}
        \centering
        \includegraphics[width=\textwidth]{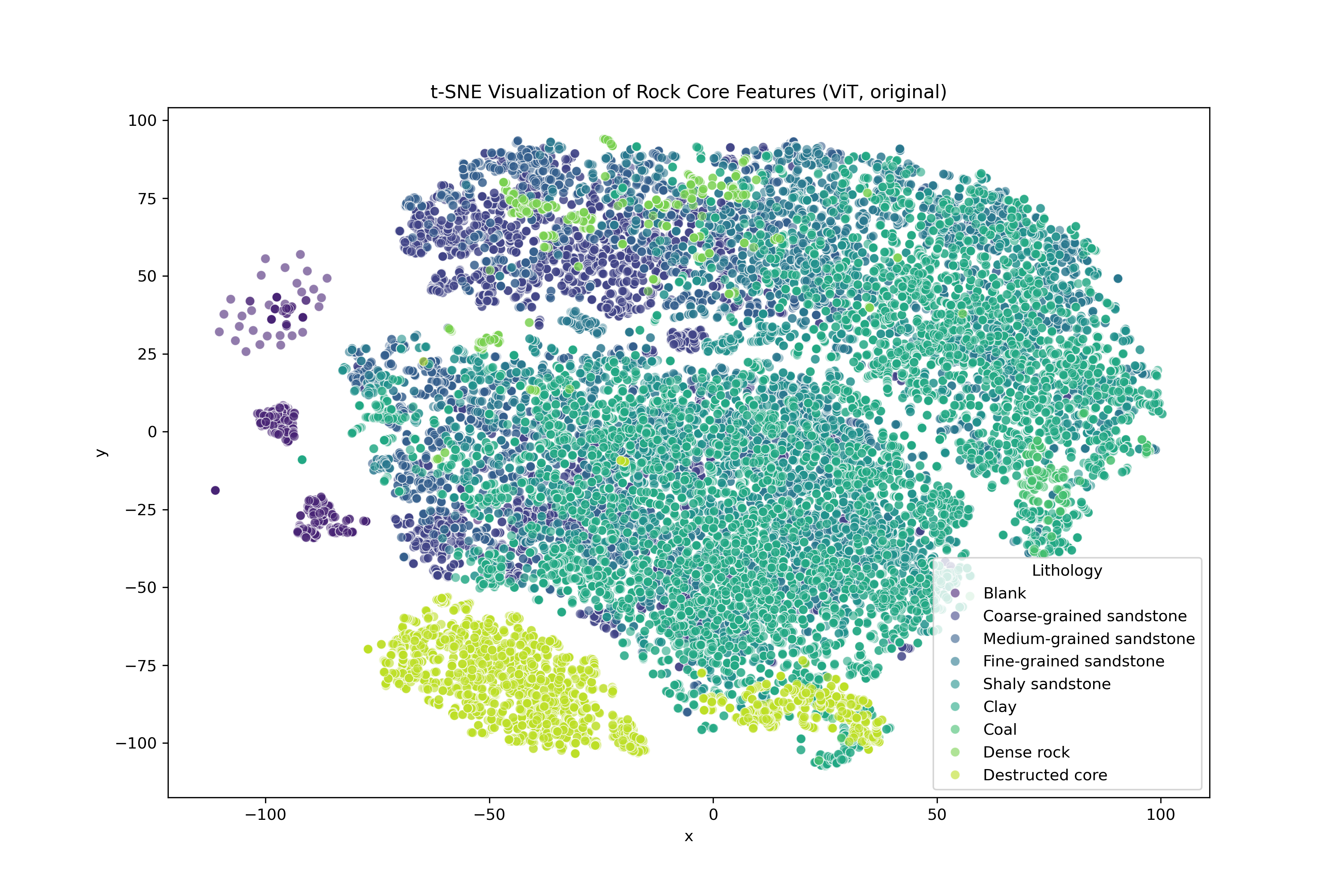}
        \caption*{(b) Inpainted Images.}
    \end{minipage}
    \caption{t-SNE visualization of high-dimensional ViT features for (a) original and (b) GAN-reconstructed core samples. The preservation of cluster topology indicates semantic consistency.}
    \label{fig:tsne_comparison}
\end{figure}

For classifiers, these hallucinated textures act as structural noise: a void in 
\textit{sandstone} filled with a generic grainy texture resembling 
\textit{siltstone} blurs decision boundaries. The t-SNE projection of ViT 
features (Fig.~\ref{fig:tsne_comparison}) confirmed that significant inter-class 
overlap existed in the original data and was preserved — not substantially 
worsened — after inpainting. The classification bottleneck was therefore driven 
primarily by pre-existing textural similarity rather than by GAN-induced 
artifacts alone.

This interaction between class overlap and generative smoothing was further 
reflected in the persistent gap between overall Accuracy and MCA across all 
architectures (Table~\ref{tab:model_comparison}): models defaulted to the 
dominant \textit{clay} class, depressing recall for minority facies.

\subsection{Frequency-Domain vs. Spatial-Domain Trade-offs}
While the spatial-domain approach successfully reconstructed macroscopic structures, the tendency to smooth high-frequency micro-textures remained a persistent limitation. Frequency-domain methodologies, such as wavelet-based generative models, offered a compelling alternative but introduced their own challenges. In geological applications, wavelet approaches could inadvertently introduce ringing artifacts and amplify sensor noise within heterogeneous core samples. Furthermore, the computational overhead of forward and inverse transforms must be balanced against the requirement for near-real-time inference in field environments.

\subsection{Dataset Constraints and Methodological Safeguards}
A primary constraint in this study, characteristic of the industrial oil and gas sector, was the limited availability of labeled core samples, resulting in inherent class imbalance. To ensure the scientific validity of our conclusions despite these constraints, two critical safeguards were implemented:

\begin{itemize}
    \item \textbf{Class-Weighting:} To counteract the dominance of specific facies, class-specific weights were applied during training. This penalized the loss function more heavily for misclassifications of minority classes, preventing the models from converging on biased solutions that ignore rare but geologically significant textures.
    \item \textbf{Bootstrapping for Statistical Robustness:} To verify that performance metrics were not artifacts of a specific data split, a bootstrap resampling technique was employed (1,000 iterations). The tight confidence intervals (e.g., $\pm$ 0.66\% for Swin Tiny) confirmed that the results were statistically stable.
\end{itemize}

These findings suggested that traditional inpainting metrics (PSNR, SSIM, FID) were insufficient predictors of utility for automated geological analysis, necessitating the development of future semantic-aware quality metrics.

\subsection{Computational Efficiency and Deployment Implications}
\label{sec:inference_time}

The inference analysis (Table~\ref{tab:inference_complexity}) revealed a notable 
disconnect between theoretical complexity and measured latency. EfficientNetV2 
achieved both the lowest complexity (2.8~GFLOPs) and latency (1.59~s per meter). 
Swin Tiny, despite moderate complexity (4.5~GFLOPs), incurred the highest latency 
(4.81~s per meter), whereas ViT — the most complex model (17.5~GFLOPs) — achieved 
a competitive 2.04~s per meter.

This FLOPs-latency gap reflects differences in operator efficiency. Swin Tiny's 
hierarchical window-based attention requires frequent data reshaping and 
partitioning, which introduces memory overhead on GPUs. Standard ViT, by contrast, 
relies on large-scale GEMM operations that are highly optimized for parallel 
execution on CUDA cores.

These computational characteristics inform the deployment recommendations in the 
following subsection.

\begin{table}[ht]
\centering
\caption{Computational complexity and inference time comparison.}
\label{tab:inference_complexity}
\resizebox{\columnwidth}{!}{
\begin{tabular}{lccc}
\toprule
\textbf{Model} & \textbf{Complexity (GFLOPs)} & \textbf{Latency/Patch (ms)} & \textbf{Total/Meter (s)} \\
\midrule
Swin-Tiny       & 4.5  & 130  & 4.81 \\
ViT             & 17.5 & 55   & 2.04 \\
EfficientNetV2  & 2.8  & 43   & 1.59 \\
ConvNeXt        & 4.4  & 70   & 2.59 \\
\bottomrule
\end{tabular}
}
\end{table}

\subsection{Deployment Scenarios and Model Selection Guidelines}
\label{sec:deployment}
The experimental results indicated that no single architecture constituted  a 
universally optimal solution for automated lithofacies classification; rather, 
the appropriate model selection was governed by the operational context.  In both cases, because cross-well overall accuracy remains in the 49--53\% range,
these configurations are intended to assist and accelerate expert geological
interpretation as a screening stage rather than to replace it; final lithofacies
assignments should remain subject to expert verification.

\textbf{Scenario~1: Field Deployment (Computational Efficiency Priority).}
In mineral exploration workflows requiring near-real-time interpretation directly 
at the wellsite, EfficientNetV2 is the recommended configuration. With a 
computational complexity of 2.8~GFLOPs and an inference latency of 1.59~s per 
meter of core, it offers the lowest processing overhead among the evaluated 
architectures while maintaining a stable overall accuracy of 53\% and an MCA 
of 48\%. Its compound scaling methodology provides resistance to the severe 
overfitting observed in Transformer-based models under data-constrained 
geological conditions, making it robust across different well domains.

\textbf{Scenario~2: Laboratory Analysis (Per-Class Completeness Priority).}
In controlled laboratory environments where the misclassification of rare but 
economically significant lithofacies (e.g., coal, dense rock) carries 
disproportionate consequences, the confidence-based hybrid ensemble 
(EfficientNetV2 + Swin~Tiny) is the recommended configuration, raising MCA 
from 48\% to 58\% at the cost of a moderate reduction in overall accuracy 
(49\%), as detailed in Section~\ref{sec:ensemble}.

The choice between these two configurations should therefore be guided by the 
relative cost of misclassification: where uniform accuracy across all facies 
is paramount, EfficientNetV2 standalone is preferred; where the detection of 
minority lithofacies is geologically critical, the hybrid ensemble is indicated.

\section{Conclusion and Future Work}
\label{sec:conclusion}

This study presented an integrated pipeline for inpainting and classification of 
damaged drill-core images and evaluated the relationship between reconstruction 
quality and downstream classification accuracy. Despite high perceptual fidelity 
(PSNR 28.7~dB, FID 74.01), classification performance remained bounded by 
inter-class textural ambiguity, confirming that standard inpainting metrics are 
insufficient predictors of semantic utility.
In light of this cross-well accuracy ceiling (49--53\%), the proposed pipeline
should be understood as a decision-support and pre-screening system that flags
candidate lithofacies and prioritizes expert attention, rather than as a fully
reliable standalone classifier. Its principal value lies in automating the
labor-intensive first pass over damaged core imagery, and in improving the
visibility of economically critical minority facies.

Two deployment configurations are proposed. For field environments, EfficientNetV2 
offers the best trade-off between accuracy (53\%) and inference speed (1.59~s per 
meter). For laboratory analysis where minority-class detection is critical, the 
hybrid ensemble (EfficientNetV2 + Swin~Tiny) raises MCA from 48\% to 58\% at the 
cost of a moderate reduction in overall accuracy (49\%).

The principal limitations of this work are the reliance on RGB imagery alone, 
which cannot resolve the spectral signatures necessary to disambiguate 
transitional lithofacies, and the constrained dataset size (2,190 images from 
a single geological deposit), which may limit cross-basin generalizability. 
Additionally, the severe class imbalance inherent to real-world geological 
data\,---\,where minority facies such as coal (0.2\%) and dense rock (1.0\%) 
remain systematically underrepresented\,---\,continues to constrain per-class 
sensitivity despite augmentation and class-weighting strategies.

Future work will address four directions: (1)~integration of hyperspectral data 
to resolve ambiguity between transitional lithofacies; (2)~uncertainty-aware 
classification that assigns lower confidence to synthetically reconstructed 
regions; (3)~development of semantic-aware inpainting quality metrics that 
correlate with classification performance; and (4)~physics-informed synthetic 
data generation for underrepresented facies to mitigate class imbalance.

\section*{Acknowledgments}

The authors would like to express their sincere gratitude to KMG Engineering for providing access to essential data, computational resources, expert consultations, and financial support, all of which were instrumental to the successful completion of this research.
\bibliographystyle{IEEEtran}

\bibliography{references}

@incollection{clark1999spectroscopy,
  title={Spectroscopy of rocks and minerals, and principles of spectroscopy},
  author={Clark, Roger N.},
  editor={Rencz, Andrew N.},
  booktitle={Manual of Remote Sensing, Volume 3: Remote Sensing for the Earth Sciences},
  edition={3rd},
  pages={3--58},
  publisher={John Wiley \& Sons},
  year={1999}
}

@article{pan2014electron,
  title={Electron paramagnetic resonance spectroscopy: Basic principles, experimental techniques and applications to {Earth} and planetary sciences},
  author={Pan, Yuanming and Nilges, Mark J.},
  journal={Reviews in Mineralogy and Geochemistry},
  volume={78},
  number={1},
  pages={655--690},
  year={2014},
  publisher={Mineralogical Society of America},
  doi={10.2138/rmg.2014.78.16}
}

@article{wang2023frequency,
  title={Frequency-to-spectrum mapping {GAN} for semi-supervised hyperspectral anomaly detection},
  author={Wang, Degang and Gao, Lianru and Qu, Ying and Sun, Xu and Liao, Wenzhi},
  journal={CAAI Transactions on Intelligence Technology},
  volume={8},
  number={4},
  pages={1258--1273},
  month={12},
  year={2023},
  publisher={Wiley},
  doi={10.1049/cit2.12154}
}

@article{wang2024global,
  title={Global Feature-Injected Blind-Spot Network for Hyperspectral Anomaly Detection},
  author={Wang, Degang and Zhuang, Lina and Gao, Lianru and Sun, Xu and Zhao, Xiaobin},
  journal={IEEE Geoscience and Remote Sensing Letters},
  volume={21},
  pages={1--5},
  year={2024},
  publisher={IEEE},
  doi={10.1109/LGRS.2024.3449635}
}

@article{ying2024multi,
  title={Multi-granularity feature enhancement network for maritime ship detection},
  author={Ying, Li and Miao, Duoqian and Zhang, Zhifei and Zhang, Hongyun and Pedrycz, Witold},
  journal={CAAI Transactions on Intelligence Technology},
  volume={9},
  number={3},
  pages={649--664},
  year={2024},
  publisher={Wiley},
  doi={10.1049/cit2.12310}
}

@article{van2004analysis,
  title={Analysis of spectral absorption features in hyperspectral imagery},
  author={{van der Meer}, Freek},
  journal={International Journal of Applied Earth Observation and Geoinformation},
  volume={5},
  number={1},
  pages={55--68},
  year={2004},
  publisher={Elsevier},
  doi={10.1016/j.jag.2003.09.001}
}

@book{gandhi2016essentials,
  title     = {Essentials of Mineral Exploration and Evaluation},
  author    = {Gandhi, S.M. and Sarkar, B.C.},
  year      = {2016},
  publisher = {Elsevier},
  isbn      = {978-0-12-805329-4},
  doi       = {10.1016/C2015-0-04648-2},
  url       = {https://www.sciencedirect.com/book/monograph/9780128053294/essentials-of-mineral-exploration-and-evaluation},
  language  = {english}
}

@inproceedings{krahenbuhl2015new,
  title={A new method for obtaining detailed mineral information on individual coal particles at the size that they are used in coke making},
  author={Hapugoda, Priyanthi and Krahenbuhl, Gregoire and O'Brien, Graham and Warren, Karryn},
  booktitle={Bowen Basin Symposium 2015},
  address={Brisbane, Australia},
  month = {09},
  year={2015}
}

@article{CORINA2018664,
  title={Automatic lithology prediction from well logging using kernel density estimation},
  author={Corina, Anisa Noor and Hovda, Sigve},
  journal={Journal of Petroleum Science and Engineering},
  volume={170},
  pages={664--674},
  year={2018},
  publisher={Elsevier},
  doi={10.1016/j.petrol.2018.06.012}
}

@article{HE2019410,
  title={Using neural networks and the Markov Chain approach for facies analysis and prediction from well logs in the {Precipice Sandstone} and {Evergreen Formation}, {Surat Basin}, {Australia}},
  author={He, Jianhua and {La Croix}, Andrew D. and Wang, Jiahao and Ding, Wenlong and Underschultz, J. R.},
  journal={Marine and Petroleum Geology},
  volume={101},
  pages={410--427},
  year={2018},
  publisher={Elsevier},
  doi = {10.1016/j.marpetgeo.2018.12.022}
}

@inproceedings{zhang2017deep,
  title={Deep Learning Method for Lithology Identification from Borehole Images},
  author={Zhang, Pengyun and Sun, J. M. and Jiang, Yanjiao and Gao, J. S.},
  booktitle={79th EAGE Conference and Exhibition 2017},
  pages={1--5},
  year={2017},
  organization={European Association of Geoscientists \& Engineers},
  doi={10.3997/2214-4609.201700945}
}

@inproceedings{Caja_10,
  author={Caja, Miguel Angel and Pe{\~n}a, Andrea Carolina and Campos, Jos{\'e} Rafael and {Garc{\'i}a Diego}, Laura and Tritlla, Jordi and Bover-Arnal, Telm and Mart{\'i}n-Mart{\'i}n, Juan Diego},
  title={{Image Processing and Machine Learning Applied to Lithology Identification, Classification and Quantification of Thin Section Cutting Samples}},
  booktitle={SPE Annual Technical Conference and Exhibition},
  pages={4002--4009},
  year={2019},
  doi={10.2118/196117-MS}
}

@inproceedings{ren2015faster,
  title={Faster {R-CNN}: Towards real-time object detection with region proposal networks},
  author={Ren, Shaoqing and He, Kaiming and Girshick, Ross and Sun, Jian},
  booktitle = {Advances in Neural Information Processing Systems},
  editor = {C. Cortes and N. Lawrence and D. Lee and M. Sugiyama and R. Garnett},
  volume={28},
  pages={91--99},
  year={2015},
  publisher = {Curran Associates, Inc.},
  url = {https://proceedings.neurips.cc/paper_files/paper/2015/file/14bfa6bb14875e45bba028a21ed38046-Paper.pdf}
}

@inproceedings{yu2019free,
  title={Free-form image inpainting with gated convolution},
  author={Yu, Jiahui and Lin, Zhe and Yang, Jimei and Shen, Xiaohui and Lu, Xin and Huang, Thomas S.},
  booktitle={Proceedings of the IEEE/CVF International Conference on Computer Vision (ICCV)},
  pages={4470--4479},
  year={2019},
  doi={10.1109/ICCV.2019.00457}
}

@inproceedings{Pang_2019_CVPR,
  author={Pang, Jiangmiao and Chen, Kai and Shi, Jianping and Feng, Huajun and Ouyang, Wanli and Lin, Dahua},
  title={Libra {R-CNN}: Towards Balanced Learning for Object Detection},
  booktitle={Proceedings of the IEEE/CVF Conference on Computer Vision and Pattern Recognition (CVPR)},
  pages={821--830},
  year={2019},
  doi={10.1109/CVPR.2019.00091}
}

@inproceedings{zhang2022resnest,
  title={ResNeSt: Split-Attention Networks},
  author={Zhang, Hang and Wu, Chongruo and Zhang, Zhongyue and Zhu, Yi and Lin, Haibin and Zhang, Zhi and Sun, Yue and He, Tong and Mueller, Jonas and Manmatha, R. and Li, Mu and Smola, Alexander},
  booktitle={Proceedings of the IEEE/CVF Conference on Computer Vision and Pattern Recognition (CVPR) Workshops},
  pages={2735--2745},
  year={2022},
  doi={10.1109/CVPRW56347.2022.00309}
}

@inproceedings{DynamicRCNN,
  title={Dynamic {R-CNN}: Towards High Quality Object Detection via Dynamic Training},
  author={Zhang, Hongkai and Chang, Hong and Ma, Bingpeng and Wang, Naiyan and Chen, Xilin},
  booktitle={Proceedings of the European Conference on Computer Vision (ECCV)},
  pages={260--275},
  year={2020},
  publisher={Springer},
  doi={10.1007/978-3-030-58555-6_16}
}

@article{zhang2012exemplar,
  title={Exemplar-based image inpainting using color distribution analysis},
  author={Zhang, Qing and Lin, Jiajun},
  journal={Journal of Information Science and Engineering},
  volume={28},
  number={4},
  pages={641--654},
  year={2012},
  publisher={Institute of Information Science, Academia Sinica}
}

@article{7987733,
  author={Li, Haodong and Luo, Weiqi and Huang, Jiwu},
  title={Localization of Diffusion-Based Inpainting in Digital Images},
  journal={IEEE Transactions on Information Forensics and Security},
  volume={12},
  number={12},
  pages={3050--3064},
  year={2017},
  publisher={IEEE},
  doi={10.1109/TIFS.2017.2730822}
}

@book{goodfellow2016deep,
  author = {Goodfellow, Ian and Bengio, Yoshua and Courville, Aaron},
  description = {Deep Learning - Ian Goodfellow, Yoshua Bengio, Aaron Courville - Google Books;
Complete MIT book available online in chapters. Book has been recommended to me by several people. I therefore recommend it to you.},
  note = {\url{http://www.deeplearningbook.org}},
  publisher = {MIT Press},
  title = {Deep Learning},
  year = 2016
}

@article{lecun2015deep,
  title={Deep learning},
  author={LeCun, Yann and Bengio, Yoshua and Hinton, Geoffrey},
  journal={Nature},
  volume={521},
  number={7553},
  pages={436--444},
  year={2015},
  publisher={Nature Publishing Group},
  doi={10.1038/nature14539}
}

@article{10.5382Geo-and-Mining-09,
  author={Orpen, John and Orpen, David},
  title={{Error-Proofing Diamond Drilling and Drill Core Measurements}},
  journal={SEG Discovery},
  number={123},
  pages={23--34},
  year={2020},
  publisher={Society of Economic Geologists},
  doi={10.5382/Geo-and-Mining-09}
}

@incollection{10.5382rev.21.07,
  author={Kramer Bernhard, Julia and Barnett, Wayne and Uken, Ron and Myers, Russell},
  title={{Chapter 7: Structural Analysis of Drill Core for Mineral Exploration and Mining: Review and Workflow Toward Domain-Based 3-D Interpretation}},
  booktitle={Applied Structural Geology of Ore-Forming Hydrothermal Systems},
  series={Reviews in Economic Geology},
  volume={21},
  pages={157--189},
  publisher={Society of Economic Geologists},
  year={2020},
  doi={10.5382/rev.21.07}
}

@article{202Editorial3,
  title={Editorial for Special Issue {3D/4D Geological Modeling for Mineral Exploration}},
  author={Wang, Gongwen},
  journal={Minerals},
  volume={13},
  number={2},
  pages={198},
  year={2023},
  publisher={MDPI},
  doi={10.3390/min13020198}
}

@article{Resolution2021,
  title={Resolution Enhancement for Drill-Core Hyperspectral Mineral Mapping},
  author={Contreras Acosta, Isabel Cecilia and Khodadadzadeh, Mahdi and Gloaguen, Richard},
  journal={Remote Sensing},
  volume={13},
  number={12},
  pages={2296},
  year={2021},
  publisher={MDPI},
  url={https://www.mdpi.com/2072-4292/13/12/2296},
  doi={10.3390/rs13122296}
}

@inproceedings{tan2020efficientdet,
  title={{EfficientDet}: Scalable and efficient object detection},
  author={Tan, Mingxing and Pang, Ruoming and Le, Quoc V.},
  booktitle={Proceedings of the IEEE/CVF Conference on Computer Vision and Pattern Recognition (CVPR)},
  pages={10778--10787},
  year={2020},
  doi={10.1109/CVPR42600.2020.01079}
}

@incollection{NEURIPS2019_9015,
  title={{PyTorch}: An Imperative Style, High-Performance Deep Learning Library},
  author={Paszke, Adam and Gross, Sam and Massa, Francisco and Lerer, Adam and Bradbury, James and Chanan, Gregory and Killeen, Trevor and Lin, Zeming and Gimelshein, Natalia and Antiga, Luca and others},
  booktitle={Advances in Neural Information Processing Systems 32},
  pages={8024--8035},
  year={2019},
  publisher={Curran Associates, Inc.},
  url = {https://proceedings.neurips.cc/paper_files/paper/2019/file/bdbca288fee7f92f2bfa9f7012727740-Paper.pdf}
}

@article{abadi2016tensorflow,
  title={{TensorFlow}: Large-Scale Machine Learning on Heterogeneous Distributed Systems},
  author={Abadi, Mart{\'\i}n and Agarwal, Ashish and Barham, Paul and Brevdo, Eugene and Chen, Zhifeng and Citro, Craig and Corrado, Greg S. and Davis, Andy and Dean, Jeffrey and Devin, Matthieu and others},
  journal={arXiv preprint arXiv:1603.04467},
  year={2016},
  doi = {10.48550/arXiv.1603.04467}
}

@article{mao2021deep,
  title={Deep residual pooling network for texture recognition},
  author={Mao, Shangbo and Rajan, Deepu and Chia, Liang-Tien},
  journal={Pattern Recognition},
  volume={112},
  pages={107817},
  year={2021},
  publisher={Elsevier},
  doi={10.1016/j.patcog.2021.107817}
}

@inproceedings{xue2018deep,
  title={Deep texture manifold for ground terrain recognition},
  author={Xue, Jia and Zhang, Hang and Dana, Kristin},
  booktitle={Proceedings of the IEEE Conference on Computer Vision and Pattern Recognition (CVPR)},
  pages={558--567},
  year={2018},
  url={https://arxiv.org/abs/1803.10896},
}

@inproceedings{cimpoi2015deep,
  title={Deep filter banks for texture recognition and segmentation},
  author={Cimpoi, Mircea and Maji, Subhransu and Vedaldi, Andrea},
  booktitle={Proceedings of the IEEE Conference on Computer Vision and Pattern Recognition (CVPR)},
  pages={3828--3836},
  year={2015},
  doi={10.1109/CVPR.2015.7299007}
}

@inproceedings{Lin_2015_ICCV,
  author={Lin, Tsung-Yu and RoyChowdhury, Aruni and Maji, Subhransu},
  title={Bilinear {CNN} Models for Fine-Grained Visual Recognition},
  booktitle={Proceedings of the IEEE International Conference on Computer Vision (ICCV)},
  pages={1449--1457},
  year={2015},
  doi={10.1109/ICCV.2015.170}
}

@inproceedings{isola2017Image,
  author={Isola, Phillip and Zhu, Jun-Yan and Zhou, Tinghui and Efros, Alexei A.},
  title={Image-to-Image Translation with Conditional Adversarial Networks},
  booktitle={Proceedings of the IEEE Conference on Computer Vision and Pattern Recognition (CVPR)},
  pages={1125--1134},
  year={2017},
  doi={10.1109/CVPR.2017.125}
}

@inproceedings{dutta2019vgg, 
    series={MM ’19},
   title={The VIA Annotation Software for Images, Audio and Video},
   url={http://dx.doi.org/10.1145/3343031.3350535},
   DOI={10.1145/3343031.3350535},
   booktitle={Proceedings of the 27th ACM International Conference on Multimedia},
   publisher={ACM},
   author={Dutta, Abhishek and Zisserman, Andrew},
   year={2019},
   month=Oct, pages={2276–2279},
   collection={MM ’19} 
}

@inproceedings{he2017mask,
  title={Mask {R-CNN}},
  author={He, Kaiming and Gkioxari, Georgia and Doll{\'a}r, Piotr and Girshick, Ross},
  booktitle={Proceedings of the IEEE International Conference on Computer Vision (ICCV)},
  pages={2961--2969},
  year={2017},
  doi={10.1109/ICCV.2017.322}
}

@article{Liko2024Augmentation,
  author={Lik{\'o}, Szil{\'a}rd Bal{\'a}zs and Holb, Imre J. and Ol{\'a}h, Viktor and Burai, P{\'e}ter and Szab{\'o}, Szil{\'a}rd},
  title={Deep learning-based training data augmentation combined with post-classification improves the classification accuracy for dominant and scattered invasive forest tree species},
  journal={Remote Sensing in Ecology and Conservation},
  volume={10},
  number={2},
  pages={203--219},
  year={2024},
  publisher={Wiley},
  doi={10.1002/rse2.365}
}

@article{gunther2025machine,
  title={Machine learning for drill core image analysis: A review},
  author={G{\"u}nther, Christian and Sim{\'a}n, Filip and Mokayed, Hamam and Liwicki, Marcus and Jansson, Nils and McDonnell, Paul and Hermansson, Tobias and Simistira Liwicki, Foteini},
  journal={Ore Geology Reviews},
  volume={187},
  pages={106974},
  year={2025},
  publisher={Elsevier},
  doi={10.1016/j.oregeorev.2025.106974},
  url={https://doi.org/10.1016/j.oregeorev.2025.106974} 
}

@article{fu2022deep,
  title={Deep learning based lithology classification of drill core images},
  author={Fu, Dong and Su, Chao and Wang, Wenjun and Yuan, Rongyao},
  journal={PLOS ONE},
  volume={17},
  number={7},
  pages={e0270826},
  year={2022},
  publisher={Public Library of Science},
  doi={10.1371/journal.pone.0270826}
}

@article{boiger2024direct,
  title={Direct mineral content prediction from drill core images via transfer learning},
  author={Boiger, Romana and Churakov, Sergey V. and Ballester Llagaria, Ignacio and Kosakowski, Georg and W{\"u}st, Raphael and Prasianakis, Nikolaos I.},
  journal={Swiss Journal of Geosciences},
  volume={117},
  number={1},
  pages={8},
  year={2024},
  publisher={Springer},
  doi={10.1186/s00015-024-00458-3}
}

\begin{IEEEbiography}[{\includegraphics[width=1in,height=1.25in,clip,keepaspectratio]{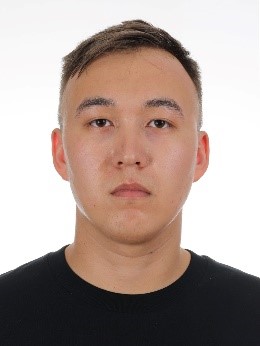}}]{G. Abdimanap}
holds a bachelor’s (2019) and a master’s (2022) degree in engineering and is currently pursuing a PhD in machine learning and data science. His research interests encompass artificial intelligence, computer vision, and ecology.
\end{IEEEbiography}

\begin{IEEEbiography}[{\includegraphics[width=1in,height=1.25in,clip,keepaspectratio]{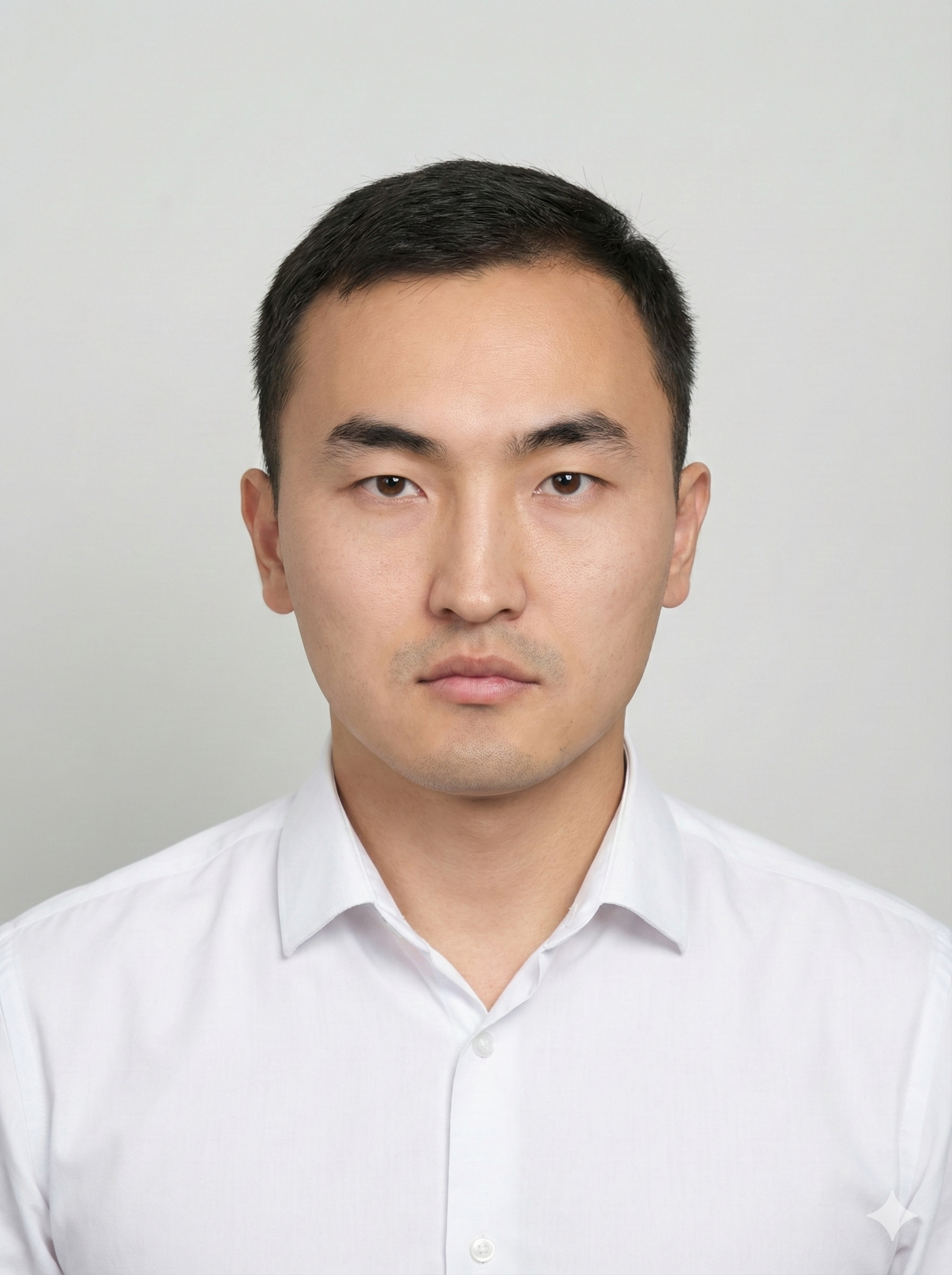}}]{K. Bostanbekov}
received the Ph.D. degree in Computer Science from the International Information Technology University (IITU), Almaty, Kazakhstan. He has contributed to the academic community through his work on various research projects involving machine learning. His current research interests focus on computer vision, modeling, and the development of intelligent systems for industrial automation.
\end{IEEEbiography}

\begin{IEEEbiography}[{\includegraphics[width=1in,height=1.25in,clip,keepaspectratio]{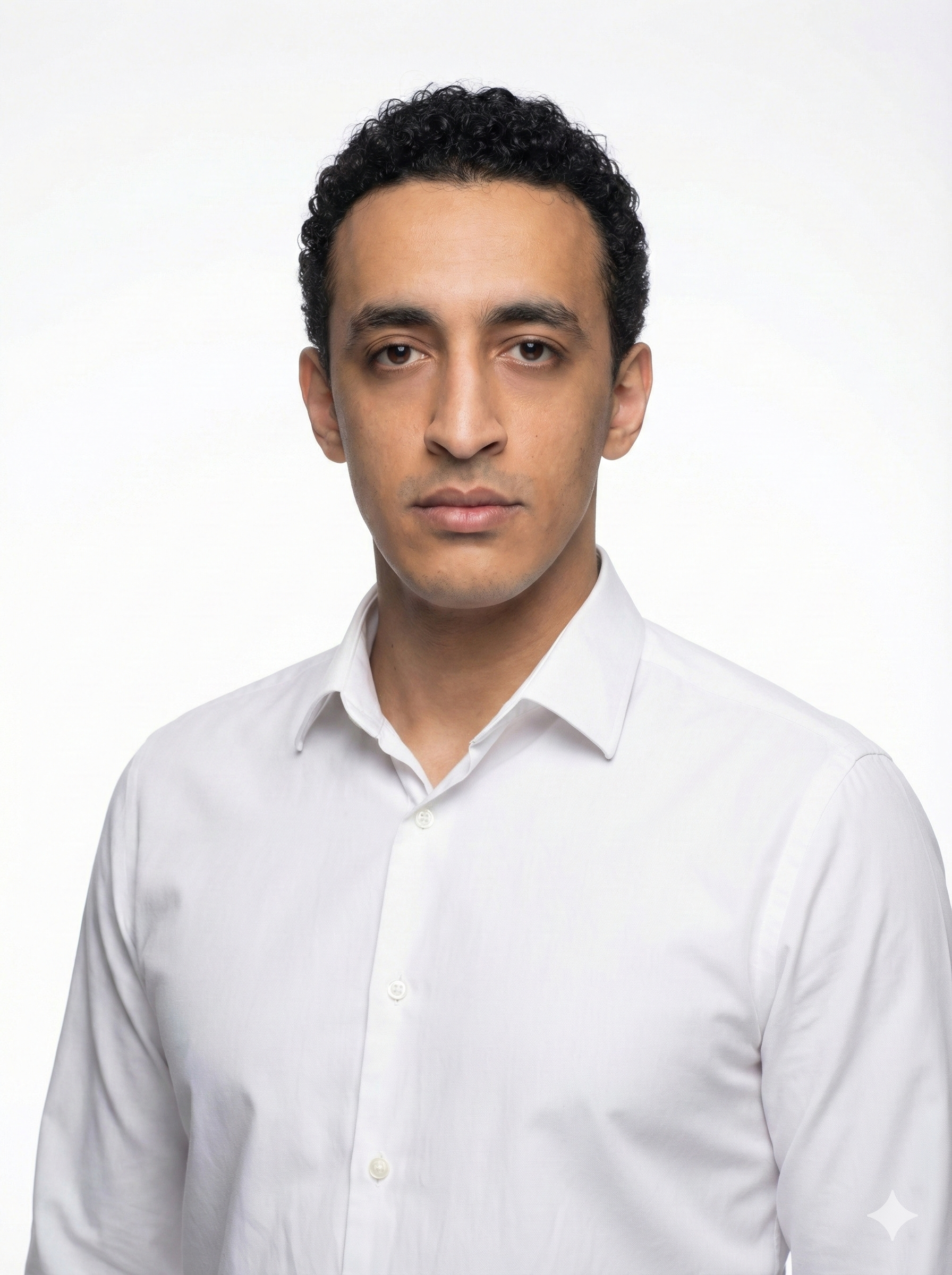}}]{A. Abdallah}
received a master's degree from Satbaev University, Almaty, Kazakhstan. He is currently pursuing his PhD at the University of Innsbruck, Austria, and holds the position of Research Assistant at the Department of Information Technology at the University of Innsbruck. His research covers several areas, including deep learning, natural language processing, and computer vision, with a particular focus on reliable learning under uncertainty.
\end{IEEEbiography}

\begin{IEEEbiography}[{\includegraphics[width=1in,height=1.25in,clip,keepaspectratio]{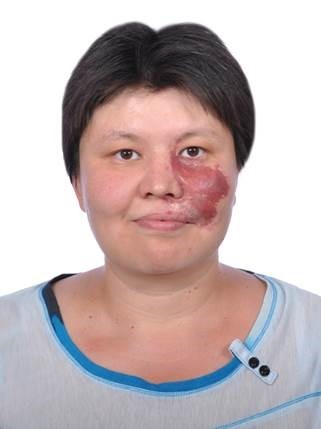}}]{A. Alimova}
graduated from the Kazakh National Pedagogical University named after Abai, receiving a master’s degree in Computer Science in 2008. In 2014, she obtained a PhD in Computational Mathematics. Since 2023, she has been working as a sector head at KMG Engineering LLP. Her professional activities and research interests cover information technology, computational mathematics, machine learning, and computer vision. She is the author of scientific publications and holds several author’s certificates.
\end{IEEEbiography}

\begin{IEEEbiography}[{\includegraphics[width=1in,height=1.25in,clip,keepaspectratio]{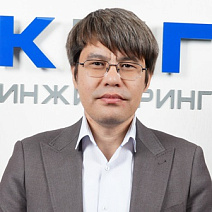}}]{D. Kurmangaliyev}
received a bachelor’s degree in Geology and Exploration of Mineral Resources from the Kazakh National Technical University named after K. I. Satpayev (KazNTU), Almaty, Kazakhstan, in 2009. He has more than 10 years of experience in geophysics, including well logging operations and geophysical data interpretation. Since 2019, he has been working at KMG Engineering, where he specializes in engineering-geophysical support and data analysis for oil and gas field development.
\end{IEEEbiography}

\begin{IEEEbiography}[{\includegraphics[width=1in,height=1.25in,clip,keepaspectratio]{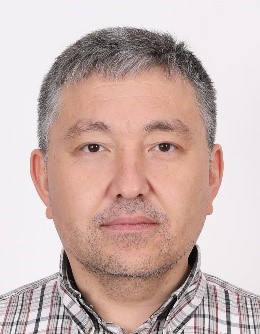}}]{D. Nurseitov}
received his Bachelor’s degree in 2001 and his Master’s degree in Physics in 2003 from Novosibirsk State University, Russia, and earned his PhD in Physical and Mathematical Sciences in 2009 in Almaty, Kazakhstan. He has experience in teaching and developing IT solutions. From 2011 to 2020, he headed the National Scientific Laboratory for Collective Use at Satpayev University, conducting grant research and technical projects. He is currently an expert advisor on the application of machine learning and computer vision methods at KMG Engineering LLP, Kazakhstan. He is the author of numerous scientific publications and articles.
\end{IEEEbiography}

\begin{IEEEbiography}[{\includegraphics[width=1in,height=1.25in,clip,keepaspectratio]{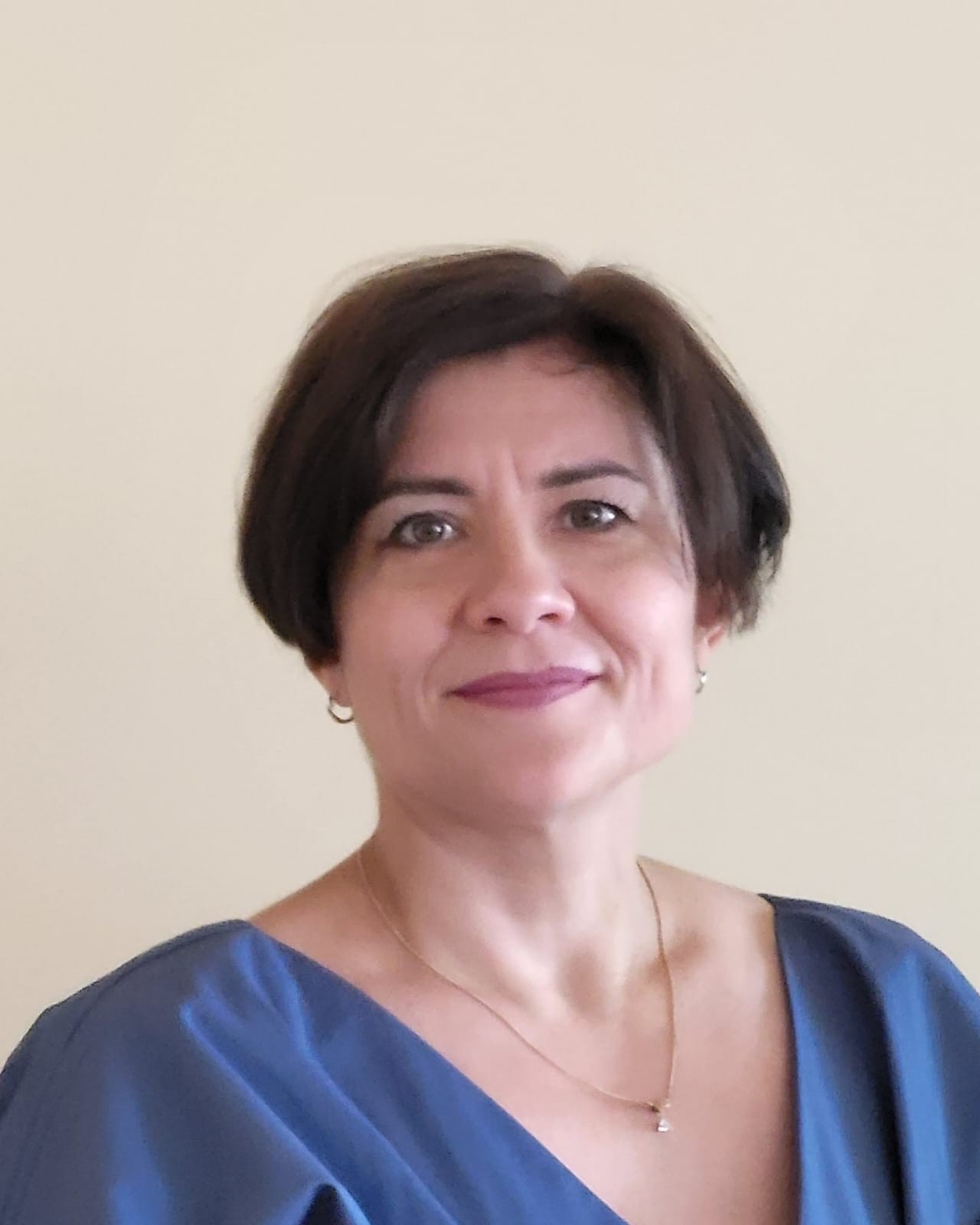}}]{T. Dedova}
 received her education at Al-Farabi Kazakh National University (Almaty) in Applied Mathematics, completed her Master’s degree, and earned her PhD at the Space Research Institute. She currently leads the Laboratory at the Institute of Ionosphere. Tatyana has participated in projects on the development of artificial neural network models for object recognition based on remote sensing data. Her research interests include environmental process modeling, environmental monitoring, and risk assessment.
\end{IEEEbiography}

\begin{IEEEbiography}[{\includegraphics[width=1in,height=1.25in,clip,keepaspectratio]{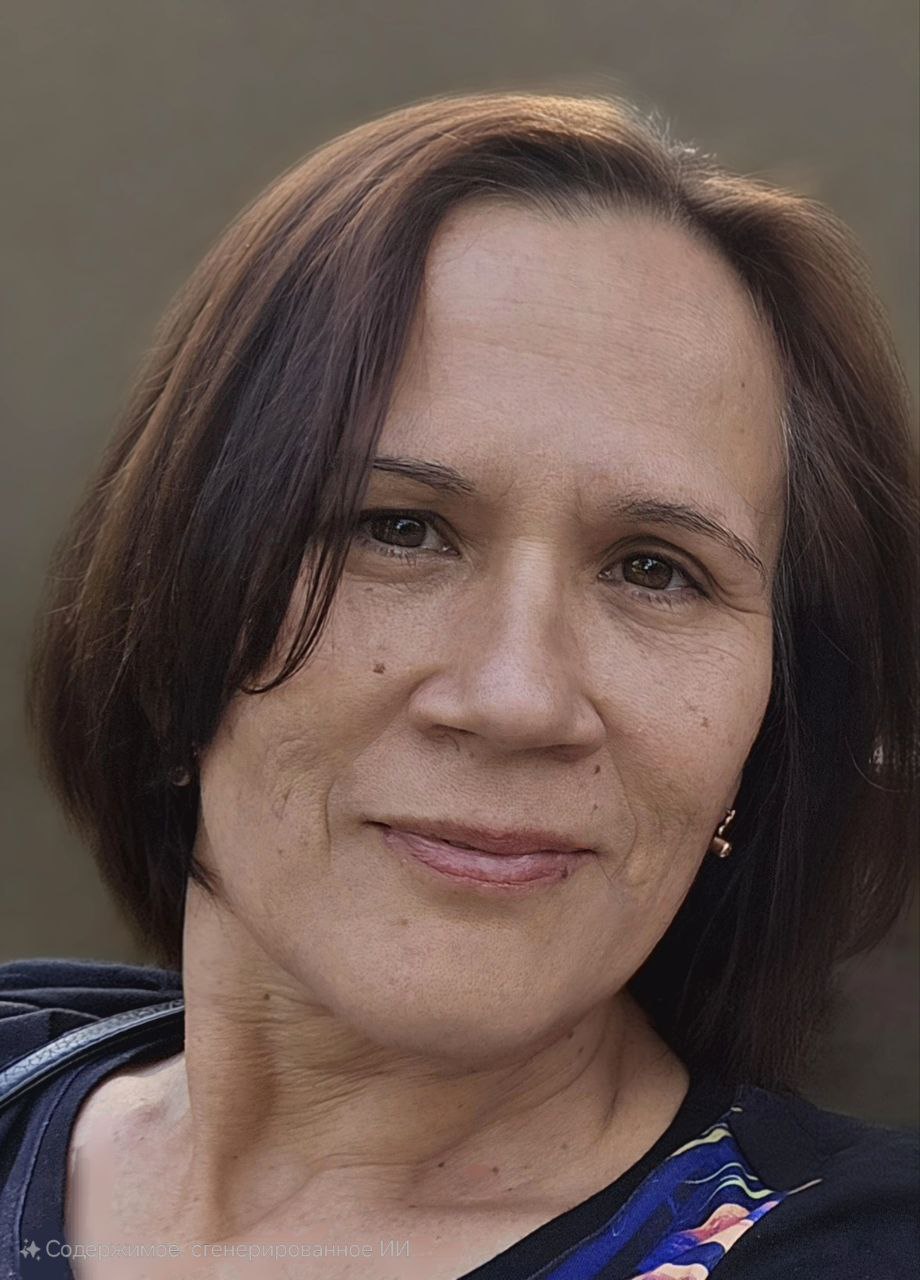}}]{L. Balakay}
graduated from Al-Farabi Kazakh National University with a degree in Applied Mathematics, where she also completed her postgraduate studies. She began her research career at the Space Research Institute and holds a PhD in Engineering. She is currently a Leading Researcher at the Institute of Ionosphere. She is actively involved in scientific projects related to hydrodynamic modeling, atmospheric monitoring, and the implementation of deep learning architectures for the analysis of remote sensing data.
\end{IEEEbiography}

\begin{IEEEbiography}[{\includegraphics[width=1in,height=1.25in,clip,keepaspectratio]{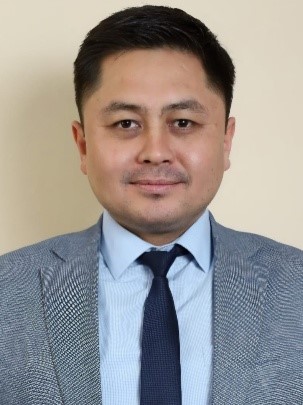}}]{S. Nurakynov}
received his MSc in Cartography from Al-Farabi Kazakh National University and his PhD in Geodesy from Satbayev University (Almaty, Kazakhstan). Since 2009, he has been affiliated with the Ionosphere Institute, where he currently serves as Director. He is the author of numerous scientific publications, including articles indexed in Scopus and Web of Science. His research interests include optical and SAR image processing, emergency monitoring, cryosphere studies, lithosphere–atmosphere–ionosphere coupling, and artificial intelligence methods.
\end{IEEEbiography}

\EOD

\end{document}